\documentclass[10pt,journal,compsoc]{IEEEtran}
\usepackage{multirow}
\usepackage{array}
\usepackage{color}
\usepackage{graphicx}
\usepackage{hyperref}
\usepackage[utf8x]{inputenc}

\usepackage{longtable} 
\usepackage{multicol}
\usepackage{booktabs}
\usepackage{todonotes}
\usepackage{textcomp}
\usepackage{pifont}
\usepackage{comment}

\usepackage{soul} 
\setuldepth{Berlin}

\usepackage{ragged2e}
\newcolumntype{L}[1]{>{\raggedright\let\newline\\\arraybackslash\hspace{0pt}}m{#1}}
\newcolumntype{C}[1]{>{\centering\let\newline\\\arraybackslash\hspace{0pt}}m{#1}}
\newcolumntype{R}[1]{>{\raggedleft\let\newline\\\arraybackslash\hspace{0pt}}m{#1}}

\definecolor{blue}{RGB}{35, 35, 166}

\ifCLASSOPTIONcompsoc
  \usepackage[nocompress]{cite}
\else
  \usepackage{cite}
\fi
\ifCLASSINFOpdf
\else
\fi
\begin{document}

%
\title{First Impressions: A Survey on Vision-based Apparent Personality Trait Analysis}


\twocolumn

%
%
%
%

\author{Julio C. S. Jacques Junior, Yağmur Güçlütürk, Marc P\'{e}rez, Umut Güçlü, Carlos Andujar, Xavier Bar\'{o}, Hugo Jair Escalante, Isabelle Guyon, Marcel A. J. van Gerven, Rob van Lier and Sergio Escalera
\IEEEcompsocitemizethanks{\IEEEcompsocthanksitem Julio C. S. Jacques Junior is a postdoctoral researcher at Universitat Oberta de Catalunya and research collaborator at Computer Vision Center, Spain \protect\\
E-mail: jsilveira@uoc.edu}
\IEEEcompsocitemizethanks{\IEEEcompsocthanksitem Yağmur Güçlütürk is an assistant professor at Radboud University, Donders Institute for Brain, Cognition and Behaviour, the Netherlands\protect\\
E-mail: y.gucluturk@donders.ru.nl}
\IEEEcompsocitemizethanks{\IEEEcompsocthanksitem Marc P\'{e}rez is a bachelor's student at University of Barcelona, Spain \protect\\
E-mail: marcperez1993@gmail.com }
\IEEEcompsocitemizethanks{\IEEEcompsocthanksitem Umut Güçlü is an assistant professor at Radboud University, Donders Institute for Brain, Cognition and Behaviour, the Netherlands\protect\\
E-mail: u.guclu@donders.ru.nl}
\IEEEcompsocitemizethanks{\IEEEcompsocthanksitem Carlos Andujar is an associate professor at Universitat Polit\`ecnica de Catalunya, Spain 
\protect\\
E-mail: andujar@cs.upc.edu}
\IEEEcompsocitemizethanks{\IEEEcompsocthanksitem Xavier Bar\'{o} is an associate professor at Universitat Oberta de Catalunya and Computer Vision Center, Spain\protect\\
E-mail: xbaro@uoc.edu}
\IEEEcompsocitemizethanks{\IEEEcompsocthanksitem Hugo Jair Escalante is a Research scientist at Instituto Nacional de Astrof\'isica, \'Optica y Eectr\'onica, Mexico \protect\\
E-mail: hugojair@inaoep.mx}
\IEEEcompsocitemizethanks{\IEEEcompsocthanksitem Isabelle Guyon is a professor and researcher at UPSud/INRIA Universit\'e Paris-Saclay, France, and Pr\'esident of ChaLearn, Berkeley, California\protect\\
E-mail: guyon@chalearn.org}
\IEEEcompsocitemizethanks{\IEEEcompsocthanksitem Marcel A. J. van Gerven is professor and PI at Radboud University, Donders Institute for Brain, Cognition and Behaviour, the Netherlands\protect\\
E-mail: m.vangerven@donders.ru.nl}
\IEEEcompsocitemizethanks{\IEEEcompsocthanksitem Rob van Lier is an associate professor at Radboud University, Donders Institute for Brain, Cognition and Behaviour, the Netherlands\protect\\
E-mail: r.vanlier@donders.ru.nl}
\IEEEcompsocitemizethanks{\IEEEcompsocthanksitem Sergio Escalera is an associate professor at University of Barcelona and Computer Vision Center, Spain \protect\\
E-mail: sergio@maia.ub.es}}
\IEEEtitleabstractindextext{%
\begin{abstract}
Personality analysis has been widely studied in psychology, neuropsychology, and signal processing fields, among others. From the past few years, it also became an attractive research area in visual computing. From the computational point of view, by far speech and text have been the most considered cues of information for analyzing personality. However, recently there has been an increasing interest from the computer vision community in analyzing personality from visual data. Recent computer vision approaches are able to accurately analyze human faces, body postures and behaviors, and use these information to infer \emph{apparent} personality traits. Because of the overwhelming research interest in this topic, and of the potential impact that this sort of methods could have in society, we present in this paper an up-to-date review of existing vision-based approaches for apparent personality trait recognition. We describe seminal and cutting edge works on the subject, discussing and comparing their distinctive  features and limitations. Future venues of research in the field are identified and discussed. Furthermore, aspects on the subjectivity in data labeling/evaluation, as well as current datasets and challenges organized to push the research on the field are reviewed. 
\end{abstract}

}

\maketitle

\IEEEdisplaynontitleabstractindextext

%
\IEEEpeerreviewmaketitle

\section{Introduction}


Psychologists have long studied human personality, and throughout the years different theories have been proposed to categorize, explain and understand it. According to Vinciarelli and Mohammadi~\cite{Vinciarelli_IEEETRANSACTIONONAFFECTIVECOMPUTING_2013}, the models that most effectively predict measurable aspects in the life of people are those based on \textit{traits}. Trait theory~\cite{Costa1998} is an approach based on the definition and measurement of traits, i.e., habitual patterns of behaviors, thoughts and emotions relatively stable over time. Trait models are built upon human judgments about semantic similarity and relationships between adjectives that people use to describe themselves and the others. For instance, consider most of people know the meaning of \textit{nervous}, \textit{enthusiastic}, and \textit{open-minded}. Trait psychologists build on these familiar notions, giving precise definitions, devising quantitative measures, and documenting the impact of traits on people's lives~\cite{Costa1998}.

Psychology studies, among other aspects, behaviour. Behaviour ($B$) is a function of the person ($P$) and the situation ($S$), i.e., $[B=f(P,S)]$. From a psychological point of view, most research has been conducted on the personal side of the equation ($P$), especially on individual differences in personality and cognitive traits. From a computational perspective, recent studies also started to pay attention on the situational part of the equation ($S$), with a particular interest on personality perception. Apparent personality ($A$), however, is conditioned to the observer ($O$), and could be defined as a function of $(P,S,O)$, i.e., $[A=f(P,S,O)]$. While a vast amount of psychological research on the study of cognitive processes in individuality judgment from the observer's point of view can be found in the literature~\cite{Funder:2012}, the research from a computational point of view is just at its early stages. This study revealed, among other things, that in addition to the measurements of agreement with respect to personality perception, almost no attention was given to the part of the equation associated to the observer ($O$) when apparent personality trait recognition is considered.

From the perspective of automatic personality computing, the relationship between \textit{stimuli} (everything observable people do) and the outcomes of the social perception processes (how we form impressions about others) is likely to be stable enough to be modeled in statistical terms~\cite{Vinciarelli2016}. This is a major advantage for the analysis of social perception processes because computing science, in particular machine learning, provides a wide spectrum of methods aimed at modeling statistical relationships like those observed in social perception. However, the main criticism against the use of trait models is that they are purely descriptive and do not correspond to actual characteristics of individuals, even though several decades of research have shown that the same traits appear with surprisingly regularity across a wide range of situations and cultures, suggesting that they actually correspond to psychological salient phenomena~\cite{Vinciarelli_IEEETRANSACTIONONAFFECTIVECOMPUTING_2013}.

During the past decades, different trait models have been proposed and broadly studied: the Big-Five~\cite{McCrae:1992}, Big-Two~\cite{Abele:2007}, 16PF~\cite{Raymond:1986}, among others. The model known as Big-Five or Five-Factor Model, often represented by the acronyms \textit{OCEAN}, is one of the most adopted and influential models in psychology and personality computing. It is a hierarchical organization of personality traits in terms of five basic dimensions: \textit{\textbf{O}penness to Experience} (contrasts traits such as imagination, curiosity and creativity with shallowness and imperceptiveness), \textit{\textbf{C}onscientiousness} (contrasting organization, thoroughness and reliability with traits as carelessness, negligence and unreliability), \textit{\textbf{E}xtraversion} (contrasts talkativeness, assertiveness and activity level with silence, passivity and reserve), \textit{\textbf{A}greeableness} (contrasts kindness, trust and warmth with traits such as hostility, selfishness and distrusts), and \textit{\textbf{N}euroticism} (contrasts emotional instability, anxiety and moodiness with emotional stability). 

\begin{figure}[tbp]
	\centering
	\includegraphics[height=3.2cm]{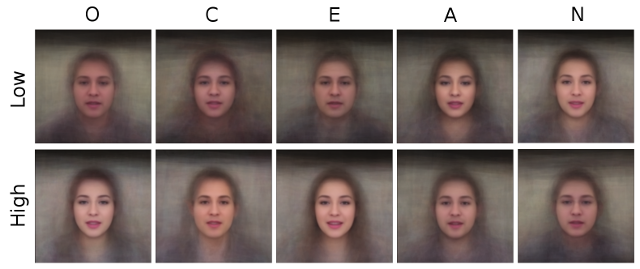}
	\caption{Representative face images for the highest and lowest levels of each Big-Five trait, obtained from~\cite{Gucluturk:TAC:2017}. Images were created by aligning and averaging the faces of 100 unique individuals that had the highest and lowest evaluations for each trait on the ChaLearn First Impression database~\cite{lopez2016chalearn} (we inverted images for \textit{Neuroticism} trait, i.e., low $\rightarrow$ high, as they were drawn in~\cite{Gucluturk:TAC:2017} from the perspective of \textit{Emotion Stability}).}
	\label{bigfive:fig}
\end{figure}

For the sake of illustration, Fig.~\ref{bigfive:fig} shows representative face images for the highest and lowest levels of each Big-Five trait, obtained from~\cite{Gucluturk:TAC:2017}. Such images reflect a kind of relationship between observers perception and observed people on the ChaLearn First Impression Database~\cite{lopez2016chalearn}. As it can be seen, these representative samples are influenced by, among other things, facial expression (e.g., a high score on \textit{Extraversion} trait and an associated smiling expression) and subjective bias of annotators with respect to gender (i.e., some traits scored high, such as \textit{Opennes to Experience} and \textit{Extraversion}, show a female looking face whereas the same traits scored low seems to show a male looking face). These are just few examples to show how the characteristics of observers and observed people affect first impressions of personality and how complex and subjective it can be. 

Assessing the personality of an individual means to measure how well the adjectives mentioned above describe him/her. Psychologists have developed reliable and useful methodologies for assessing personality traits. Despite its known limitations, the self-report questionnaires have become the dominant method for assessing personality~\cite{Gregory:2009}. However, personality assessment is not limited to psychologists: everybody, everyday, makes judgments about our personalities as well as of others. In every-day intuition, the personality of a person is assessed along several dimensions. We are used to talk about an individual as being too much/little focused on herself, (dis-)organized, (non-)open-minded, etc~\cite{Pianesi:2008:MRP:1452392.1452404}. Nonetheless, support for the validity of these first impressions is inconclusive, raising the question of \textit{why do we form them so readily?} According to Willis and Todorov~\cite{Willis:2006}, people make first impressions about others, either from their faces~\cite{Todorov:2017:book} or in general, from a glimpse as brief as 100ms, and these snap judgments predict all kinds of important decisions (discussed in Sec.~\ref{firstimpressions:intro}).

While being diverse in terms of data, technologies and methodologies, all domains concerned with personality computing consider the same three main problems~\cite{Vinciarelli_IEEETRANSACTIONONAFFECTIVECOMPUTING_2013}, namely automatic personality recognition (i.e., the \textit{real} personality of an individual), automatic personality perception (the prediction of the personality others attribute to a given person), and automatic personality synthesis (i.e., the generation of artificial personalities through embodies agents). Recently, the machine learning research community adopted the terms of \textit{apparent personality}, \textit{personality impressions}, or simply \textit{first impressions}~\cite{lopez2016chalearn,NJULambda,guccluturk2016deep} to refer to personality perception (which are used interchangeably along the text). Although, first impressions are not restricted to personality. In general, automatic personality perception is composed of three main steps: data annotation/ground-truth generation, feature extraction and classification/regression, which will be incrementally discussed along the text.

Vinciarelli and Mohammadi~\cite{Vinciarelli_IEEETRANSACTIONONAFFECTIVECOMPUTING_2013} presented the first survey on personality computing. However, they focused on automatic personality recognition, perception and synthesis from a more general point of view rather than from a vision-based perspective. In this work, we contribute to the research area in the following directions:

\begin{itemize}
\item We present an up-to-date literature review on apparent personality trait analysis from a vision-based perspective, i.e.,  centered on the visual analysis of humans. Reviewed works include some kind of image-based analysis at some stage of their pipelines. Hence, this study can be considered the first comprehensive review covering this particular research area.
\item We discuss the subjectivity in data labeling (and evaluation protocols) from first impressions, which is a relatively new and emerging research topic.
\item We propose a taxonomy to group works according to the type of data they use: \textit{still images}, \textit{image sequences}, \textit{audiovisual} or \textit{multimodal}. We claim the type of data and application are strongly correlated, and that the proposed taxonomy can help future researchers in identifying: 1) common features, databases and protocols employed in different categories, 2) pros and cons of ``similar'' approaches, and 3) what methods they should compare with (or get inspired by).
\item We present and discuss relevant works developed for \textit{real} personality trait analysis, as well as those correlating both \textit{real} and apparent personalities, which is an almost unexplored area in visual computing.
\item We present a set of mid-level cues, collected from the reviewed works, highly correlated to each personality trait of the Big-Five model. We analyse reported results and reveal commonly best recognized traits, as well as the more challenging ones.
\item We discuss current datasets and competitions organized to push the research in the field, main limitations and future research prospects. 
\item We identify recent trend methodologies, open challenges and research opportunities in the field. 
\end{itemize}

The remainder of this paper  is organized as follows.  Sec.~\ref{firstimpressions:intro} motivates the research on the topic. Subjectivity is discussed in Sec.~\ref{sec:labeling}. State-of-the-art, according to the proposed taxonomy, is presented and discussed from Sec.~\ref{sec:stillimages} to Sec.~\ref{sec:multimodal}. Then, a joint discussion about \textit{real} and apparent personality is presented in Sec.~\ref{APR:section}. Later, we briefly discuss high-level features and their correlated traits in Sec.~\ref{features:section}, and overall accuracy obtained for different Big-Five traits in Sec.~\ref{recognitionaccuracy:section}. In Sec.~\ref{challenges:section} we discuss past and current challenges organized to push the research area. Finally, final remarks and conclusions are drawn in  Sec.~\ref{conslusion:section}. 

\section{Related work} \label{relatedwork:section}

This section presents a comprehensive review on vision-based methods for apparent personality trait analysis.

\subsection{The importance of first impressions in our lives}\label{firstimpressions:intro}

A computer program capable of predicting in mere seconds the psychological profile of someone could have wide application for companies as well as for individuals around the globe. Just to mention a few, they could be applied in health (e.g. personalized psychological therapies), robotics (e.g., humanized and social robots), learning (e.g. automatic tutoring systems), leisure and business (e.g. personalized recommendation systems). For instance, recent studies show that therapeutic robots can be helpful in stimulating social skills in children with autism, 
which would be impossible without a robot provided with some advanced capabilities. Other studies indicate video interviews, through nonverbal visual human behavior analysis, are starting to modify the way in which applicants get hired~\cite{nguyen2016hirability}. Nevertheless, these kind of applications will only be truly accepted and trusted if explainability and transparency can be guaranteed~\cite{Liem2018}. Moreover, to become inclusive and benefit everyone, such systems need to be able to generalize to different contexts.

The development and evaluation of automatic methods for personality perception is a very delicate topic, making us to think over the still open question \textit{``what should be the limit of such technology?''.} The accuracy performance of apparent personality recognition models is generally measured in terms of how close the outcomes of the approach to the judgments made by external observers (i.e., annotators) are. The main assumption behind such evaluation is that social perception technologies are not expected to predict the actual state of the target, but the state observers attributed to it, i.e., their impressions. Thus, making automatic apparent personality trait analysis a very complex and subjective task. 


According to the literature, faces are rich source of cues for apparent personality attribution~\cite{biel2012facetube, Gurpinar:ECCV:2016}. However, Todorov and Porter~\cite{Todorov2014} showed that first impressions based on facial analysis can vary with different photos (i.e., the ratings vary w.r.t. context). Whether or not trait inferences from faces are accurate, they also affect important social outcomes. For example, attractive people have better mating success and jobs prospects than their less fortunate peers. The effects of appearance on social outcomes may be partly attributed to the \textit{halo effect}~\cite{Todorov:book:2012}, which is the tendency to use global evaluations to make judgments about specific traits (e.g., \textit{attractiveness} correlates with perceptions of \textit{intelligence}), having a strong influence on how people build their first impressions about others. Rapid judgments of competence based solely on the facial appearance of candidates was enough to allow participants from an experiment~\cite{Todorov:PNAS:2007} to predict the outcomes of Senate elections in the United States in 72.4\% of the cases. It seems politicians who simply look more competent are more likely to win elections~\cite{Olivola2010,Joo:ICCV:2015}. 

First impressions also influence legal decision-making~\cite{Todorov:cognition}. Nevertheless, at the same time that different research communities (e.g., machine learning, computer vision and psychology) are advancing the state-of-the-art in the field in different directions, it was recently observed\footnote{\url{https://www.theguardian.com/technology/2017/apr/13/ai-programs-exhibit-racist-and-sexist-biases-research-reveals}} that some Artificial Intelligence based models are exhibiting racial and gender biases, which are considered extremely complex and emerging issues. One possible explanation for these problems (i.e., in the case of first impressions) is that the ground truth annotations, used to train AI based models, are given by individuals and may reflect their bias/preconception towards the person in the images or videos, even though it may be unintentional and subconscious~\cite{Escalante:2018:arXiv:IJCV}. Hence, trained classifiers can inherently contain a subjective bias. This phenomena was also observed and analyzed in natural language processing~\cite{Caliskan:2017}, and should receive special attention from all research areas.


\subsection{How challenging and subjective can be apparent personality trait labeling/evaluation?}\label{sec:labeling}

The outcomes of machine learning based models will be in some way a reflect of the learning data they use, i.e., in the case of first impression analysis, the  labels provided by the annotators. The validity of such data can be very subjective due to several factors, such as  cultural~\cite{Walker:2011,Sofer:2017:Percep}, social~\cite{Sutherland:2016,Todorov:book:2012}, contextual~\cite{Todorov2014}, gender~\cite{Mattarozzi:PLOS:2015, Jenkins2011313}, appearance~\cite{Rudert2017101}, etc., which makes the research and development on personality perception a very challenging task. 

The subjectivity of impressions raises further questions about how many annotators should be involved in an experiment and how much they should agree with one another. When it comes to the Big-Five model, the literature suggests that the agreement should be measured in terms of amount of variance shared by the observers. In general, the low agreement should not be considered the result of low quality judgments or data, but an effect of the inherent ambiguity of the problem~\cite{Vinciarelli2016}. However, existing works analyse apparent personality from a universal perception, that is, the impressions given by different observers concerning a particular individual are averaged. Although this became a standard procedure, we argue it does not accurately reflect how first impressions work in real life. Person perception is conditioned to the observer. Thus, particularities of the different populations of observers, such as cultural differences~\cite{Walker:2011}, should also be taken into account. 
The aim of this section, rather than trying to answer the above questions, is to introduce a brief review and discussion on the topic.

Jenkins et al.~\cite{Jenkins2011313} analyzed the variability in photos of the same faces. According to their study, within-person variability exceeded between-person variability in \textit{attractiveness}, suggesting that the consequences of within-person variability are not confined to judgments of identity. It was also observed that female annotators tended to be rather harsh on the male faces (i.e., gender issues). Although attractiveness is not explicitly related to personality, results indicate how complex is the task of social judgment. The study presented in~\cite{Sutherland:2016} revealed that the most important source of within-person variability, related to social impressions of key traits of \textit{trustworthiness}, \textit{dominance} and \textit{attractiveness}, which index the main dimensions in theoretical models of facial impressions, is the emotional expression of the face, but the viewpoint of the photograph also affects impressions and modulates the effects of expression.

Within-person variance in behavior is likely to be a response to variability in relevant \textit{situational cues} (e.g., people are more extraverted in large groups than in small groups - even though some individuals may not increase or may even decrease their level of \textit{extraversion} with the size of the group)~\cite{Fleeson:2001}. Because situational cues vary in everyday behavior, behavior varies as well. Abele and Wojciszke~\cite{Abele:2007} analyzed the Big-Two model (i.e., \textit{Agency}, also called competence, power, and \textit{Communion}, also called warmth, morality, expressiveness) from the perspective of self versus others. 
According to their work, \textit{agency} is more desirable and important in the self-perspective, and \textit{communion} is more desirable and important in the other-perspective, i.e.,  \textit{``people perceive and evaluate themselves and others in a way that maximizes their own interests and current goals''}.

Cross cultural consensus and differences were found in~\cite{Walker:2011} when addressing the problem of universal and cultural differences in forming  personality trait judgments from faces. Using a face model, they were able to formalize the static facial information that is used to make certain personality trait judgments such as \textit{aggressiveness}, \textit{extroversion} and \textit{likeability}. According to their study, Asian and Western participants were able to identify the enhanced salience of all different traits in the faces, suggesting that the associations between mere static facial information and certain traits are highly shared among participants from different cultural backgrounds. 
Nevertheless, faces with enhanced salience of \textit{aggressiveness}, \textit{extroversion}, and \textit{trustworthiness} were better identified by Western than by Asian participants.

Even though the problem of stereotyping is minimized in~\cite{Walker:2011} (which could bias the analysis) through the usage of manipulated faces (synthetic faces are used in~\cite{Todorov:cognition, Olivola2010} for the same purpose), it should be noted that social judgments in real situations are formed from different sources, such as pose, gaze, facial expression or styling. For example, hairstyle, which is extrafacial information, can be intentionally chosen by target persons to shape others' impressions of them. According to Vetter and Walker~\cite{MirellaWalker:2011}, in order to systematically investigate how faces are perceived, categorized or recognized, we need to control over the \textit{stimuli} we use in ours experiments. The problem is not only to get face images taken under comparable lighting conditions, distance from the camera, pose or facial expression, but to get face \textit{stimuli} with clearly defined similarities and differences.

Barratt et al.~\cite{Barratt:2016} revisited the classical problem of the \textit{Kuleshov effect}. According to film mythology, the filmmaker Lev Kuleshov conducted an experiment (in the early 1920s) in which he combined a close-up of an actor's neutral face with three different emotional contexts: happiness, sadness and hunger. The viewers reportedly perceived the actor's face as expressing an emotion congruent with the given context. Recent attempts at replicating the experiment have produced either conflicting or unreliable results. However, it was observed that some sort of \textit{Kuleshov effect} does in fact exist. Olivola and Todorov~\cite{Olivola2010315} evaluated the ability of human judges to infer the characteristics of others from their appearances. They found that judges are generally less accurate at predicting characteristics than they would be if appearance cues are ignored, suggesting that appearance is overweighed and can have detrimental effects on accuracy.

More recently, Escalante et al.~\cite{Escalante:2018:arXiv:IJCV} analyzed the \textit{First Impression} dataset~\cite{lopez2016chalearn} and top winner approaches used in the ChaLearn LAP Job Candidate Screening  Challenge~\cite{Escalante:IJCNN:2017}.
Part of the study focused on the existence of latent bias towards gender and ethnicity. 
When correlating these variables with apparent  personality annotations (Big-Five), they first observed that there was an overall positive attitude/preconception towards females in both personality traits (except for \textit{agreeableness}) and job interview invitation variable. Moreover, gender bias was observed to be stronger compared to ethnicity bias. Concerning ethnicity, results indicated an overall positive bias towards Caucasians, and a negative bias towards African-Americans. No discernible bias towards Asians in either way was observed.

\subsubsection{Discussion}

This section discusses Sec.~\ref{sec:labeling}, covering the main challenges related to data labeling, different solutions employed to address worker bias, and suggestions for future research.

\textbf{Data labels.} Annotation protocols for personality perception require special attention. The challenge relies in defining a particular score to a certain trait, either from a continuous domain or within a specific range (e.g., using Likert Scale~\cite{Biel:ICMI:2013}), which by default is extremely subjective, time-consuming and influenced by worker bias. This study revealed that there is no standard protocol to annotate data for personality perception, as well as there is no gold standard for apparent personality. According to~\cite{joshi2014automatic}, every individual may perceive others in a different way, which poses a great obstacle for creating a model addressing the issues of annotators subjectivity and bias.   
For the sake of illustration of how challenging data labeling in this area can be, Nguyen and Gatica-Perez~\cite{nguyen2016hirability} asked Amazon Mechanical Turk (AMT) workers to label $939$ videos with respect to the Big-Five model using a five-point Likert Scale and a standard personality inventory questionnaire. In their work, only \textit{extraversion} trait was observed to be consistently rated.  

\textbf{Reducing bias.} Worker bias is particularly difficult to evaluate and correct when many workers contribute just a few labels, which is typical when labeling is crowd-sourced. Bremner et al.~\cite{Bremner:HRI:2016} proposed to identify annotators who assigned labels without looking at the content by removing those who incorrectly answered a test question. Pairwise comparisons~\cite{lopez2016chalearn,Chen2016,Joo:ICCV:2015} demonstrated in recent studies to be a very effective way to address worker bias. Joo et al.~\cite{Joo:ICCV:2015} asked AMT workers to compare a pair of images in given dimensions rather than evaluating each image individually. A similar strategy is applied in~\cite{lopez2016chalearn} for video files, which include an algorithm to estimate continuous scores (i.e., ground truth labels) from pairwise annotations~\cite{Chen2016}. Comparison schemes have three main advantages: 1) the annotators do not need to establish the absolute baseline or scales in these social dimensions, which would be inconsistent, i.e., ``\textit{what does a score of $0.3$ mean?}''; 2) they naturally identify the strength of each sample in the context of relational distance from other examples, generating a more reliable ranking of subtle signal differences~\cite{Joo:ICCV:2015}; and 3) they avoid previously annotated samples from biasing future scores (i.e., scoring someone very low on a certain trait because of an unconscious comparison with previous videos/images where the score was high).

\textbf{Future directions.} The area of personality computing would benefit from the design, definition and release of new, large and public databases, as well as with the design of standard protocols for data collection, labeling and evaluation. According to our study, \texttildelow40\% of reviewed works developed for personality perception are evaluated on public databases without any type of customization (in most of the cases, small in size and/or composed of small number of participants);  around 25\% of works are evaluated on private databases, and the remaining ones on customized from public databases. The use of private or customized databases makes the comparison among works a big challenge, creating a barrier to advance the state-of-the-art on the field. Disregarding the different applications (e.g., face-to-face interviews, HRI, or conversational videos), which can influence the design of new databases, we argue that future developed databases should be composed at least of a large number of samples from a heterogeneous population (with respect to observed people and annotators), so that current and/or future works can generalize to different cultures easily. We envisage two main scenarios can make a big difference in future researches on the topic, as follows.

\ul{Joint analysis of \textit{real} and apparent personalities}: it will require the design of new, large and public databases containing labels for both \textit{real} and apparent personalities. 
Up to now, the correlation analysis of both personality types, from a computer vision/machine learning point of view, have not been fully exploited. Despite the great difficulty of accomplish such task, as it requires data collection (self-reports and annotations) from a large population, it could benefit different research lines on the field (discussed in Sec.~\ref{APR:section}). For instance, when addressing real and apparent age estimation, Jacques et al.~\cite{Jacques:FG:2019} show that the subjective bias contained in the perception of age can be used to better approximate the real target. As far as we know, a similar idea has never been exploited in the context of personality trait analysis. Then, questions such as \textit{``what is the effect of the real personality of someone on his/her perceived personality?''}, or \textit{``is it possible to accurately regress the real personality from perception mechanisms?''}, could be studied.

\ul{Correlate observer \textit{vs.} observed}: \textit{``what is the observer looking at?''} or \textit{``what characteristics does the observer have?''}, or even better, a combination of both questions. Existing works are not considering any information about the observers (apart from their impressions) to perform automatic personality perception, such as cultural similarities or differences~\cite{Walker:2011} with respect to different target populations. Future researches could take into account, for example, the gender, age, ethnicity, or even the \textit{real} personality (making a link to the previous scenario) from both observers and people being observed. Taking the observers characteristics into account would move the research on this area to another level. Nevertheless, the above questions may pose a great challenge to privacy and ethic issues. A preliminary analysis on this topic was addressed in~\cite{Escalante:2018:arXiv:IJCV}. Authors analyzed the correlation among gender and ethnicity (from the people being observed) \textit{vs.} the interview variable contained in their database (provided by the observers). However, any data from the annotators were collected.

Some extra questions still remain open and could be a subject for future researches, such as whether universal and cultural differences studied in~\cite{Walker:2011} can be generalized to faces from other cultural backgrounds, or whether the personality impression about one trait can influence the impressions about other traits~\cite{teijeiro2015your}, or the relationship among nonverbal content and personality variables/scores~\cite{Naim:FG:2015}.

\subsection{Still Images}\label{sec:stillimages}
This section reviews the very few works developed for automatic personality perception from still images (neither audio nor temporal information are used). This class of works usually focus on facial information to drive their models, generally combining features at different levels and their relationships. Note that some works developed for other categories (e.g., image sequences or audio-visual) could be applied (or easily adapted) to still images, as they perform a frame-by-frame prediction before a final aggregation or fusion (responsible to deal with the temporal information).

In the work of Guntuku et al.~\cite{Guntuku:2015}, low-level features are employed to detect mid-level cues (gender, age, presence of image editing, etc), used to predict \textit{real} and apparent personality traits (Big-Five) of users in self-portrait images (\textit{selfies}). Even though a small dataset is used (i.e., composed of 123 images from different users), authors presented some insights on which mid-level cues contribute to personality recognition and perception (see Table~\ref{tab:feat:correlation}). Yan et al.~\cite{Yan:MMM:2016} studied the relationship between facial appearance and personality impressions in the form of \textit{trustworthy}. In their work, different low-level features are extracted from different face regions, as well as relationships between regions. For instance, HoG is used to describe eyebrow shape, and Euclidean distance to describe eyes width. To alleviate the semantic gap between low-level and high-level features, mid-level cues are built through clustering. Then, a Support Vector Machine (SVM) is used to find the relationship between face features and personality impression.

Dhall and Hoey~\cite{Dhall:HBU:2016} exploited a multivariate regression approach to infer personality impressions of users from  Twitter profile images. Hand-crafted and deep learning based features are computed from the face region. Background information is considered, which may affect personality perception, and a high correlation between \textit{openness} and scene descriptors was observed, suggesting that the context where pictures are taken loosely relate to a person's ability to explore new places. In~\cite{AlMoubayed:2014:FAP:2647868.2655014}, authors combined \textit{eigenfaces} features with SVM to investigate whether people perceive an individual portrayed in a picture to be above or below the median with respect to each Big-Five trait.

\subsubsection{Discussion}
This section discusses the importance of semantic attributes, limitations and future research directions related to Sec.~\ref{sec:stillimages}.

\textbf{Mid-level features.} Facial landmarks seem to be the start point of different feature extraction methods~\cite{Dhall:HBU:2016}, especially those which exploit mid-level features or semantic attributes~\cite{Guntuku:2015,Yan:MMM:2016}. Mid-level features or semantic attributes usually carry meaningful information, which can be used to complement other low-level features and then improve accuracy performance. They also enable more interpretable analysis on the results. For instance, when studying social impressions, Joo et al.~\cite{Joo:ICCV:2015} analyzed the correlation between a set of mid-level attributes and social dimensions (e.g., \textit{attractiveness}, \textit{intelligence} and \textit{dominance}), and investigated which face regions contribute more to each trait dimension. Such analysis could be considered a step forward in the direction of feature selection for automatic personality perception. Moreover, mid-level cue detectors outperformed many low-level features analysed in~\cite{Guntuku:2015}, for almost all trait dimensions of the Big-Five model, reinforcing their benefit.

\textbf{Current limitations.} Most works presented in this section either built their own datasets~\cite{Guntuku:2015} (e.g., collecting data from the Internet) or adapted to their needs~\cite{AlMoubayed:2014:FAP:2647868.2655014,Yan:MMM:2016} datasets developed for other purposes 
(e.g., face recognition). The common point of these works is that images need to be labelled, and labels usually do not become public. This way, reproducing their results can be a big challenge.  Twitter profile images collected in~\cite{Liu:AAAI:2016} are used in~\cite{Dhall:HBU:2016}. However, baseline labels (Big-Five) are created trough the analysis of users tweets. These points reinforce the fact that new and large public datasets, with associated standard protocols, are fundamental to advance the state-of-the-art on the field.

\textbf{Future directions.} The analysis of image content outside the face region, as performed in~\cite{Dhall:HBU:2016}, is a topic which deserves further attention (and not specifically related to still images, as it could be applied to other categories). As emphasized in~\cite{Kosti:CVPR:2017}, when addressing the perception of emotions from images, the context have an important role, which is completely aligned with personality perception studies. Although some works proposed to ignore the background information, hairstyle or clothes~\cite{Walker:2011, Todorov:cognition, Olivola2010}, people can intentionally combine such information to shape others' impressions of them. Moreover, the literature shows that context also influence annotators during data labeling. In addition to the context, high-level features extracted from body pose, gestures or facial expression, which have already been exploited by some works in other categories, have not been fully exploited when still images are considered. Body language analysis~\cite{Noroozi:arXiv2018}, which is an emerging research topic in computer vision, could benefit apparent personality trait analysis in many ways. It includes different kinds of nonverbal indicators, such as gaze direction, position of hands, the style of smiling, among others, which are important markers of the emotional and cognitive inner state of a person.

\subsection{Image Sequences}\label{sec:imagesequences}
Works exploiting visual cues of image sequences are presented next. They benefit from temporal information and scene dynamics (without acoustic information), which bring useful and complementary information to the problem.

Biel et al.~\cite{biel2012facetube} studied personality impressions in conversational videos (\textit{vlogs}) using a subset of the Youtube \textit{vlog} dataset~\cite{biel2010voices} and focusing on facial expression analysis. 
Automatic personality perception is addressed using Support Vector Regression (SVR) combined with statistics of facial activity  based on frame-by-frame estimates. 
Results show that \textit{extraversion} is the trait showing the largest activity cue utilization (reinforced in Table~\ref{tab:feat:correlation}), which is related to evidences found in the literature that \textit{extraversion} is typically easier to judge~\cite{biel2011you,biel2013youtube}. Later, Aran and Gatica-Perez~\cite{aran2013cross} investigated the use of social media content as a domain for transfer learning from conversational videos to small group settings. They considered the particular trait of \textit{extraversion}, and addressed the problem combining Ridge Regression and SVM classifiers with statistics extracted from \textit{weighted Motion Energy Images}. In~\cite{teijeiro2015your}, the connections between facial emotion expressions and personality impressions are analysed as an extension of~\cite{biel2012facetube}. Four sets of behavioral cues (and fusion strategies) that characterize face statistics and dynamics over brief observation windows are proposed to represent facial patterns. Co-occurrence analysis (e.g., smiling with surprise) is also exploited. Finally, the inference task is addressed using SVR. Their study show that while multiple facial expression cues have significant correlation with several of the Big-Five traits, they were only able to significantly predict \textit{extraversion} impressions.

Taking Human-Computer Interaction (HCI) into account, Celiktutan and Gunes~\cite{celiktutan2014continuous} addressed the challenging task of continuous prediction of perceived traits in space and time. According to the authors, continuous predictions of first impressions have not been explored before. In their work, external observers were asked to continuously provide ratings along multiple dimensions (Big-Five) in order to generate continuous annotations of video sequences. The inference problem is then addressed using low-level visual features (e.g., HoG/HoF) combined with a linear regression method. 
The work was extended in~\cite{Celiktutan:2014:MFA:2668024.2668026, Celiktutan:FG:2015} with real-time capabilities, audio-only and audio-visual data analysis.

Considering the great advances in the field of deep learning, G{\"u}rpinar and collaborators~\cite{Gurpinar:ECCV:2016} employed a pre-trained Convolutional Neural Network (CNN) to extract facial expressions as well as ambient information (which is ignored by most competitive works on the topic) on the ChaLearn First Impression~\cite{lopez2016chalearn} dataset. Visual features that represent facial expressions and scene are combined and fed to a Kernel Extreme Learning Machine (ELM) regressor. Ventura et al.~\cite{Ventura:CVPRW2017} studied why CNN models are performing surprisingly well in automatically inferring first impressions. Results show that the face provides most of the discriminative information for personality impressions inference, and the internal CNN representations mainly analyze key face regions such as the eyes, nose, and mouth.

Unlike the later works, Bekhouche et al.~\cite{Edine17} combined texture features, extracted from the face region, with five SVRs to estimate apparent personality traits (Big-Five). As reported by the authors, although deep learning-based approaches can achieve better results, temporal face texture-based approaches are still very effective.

\subsubsection{Discussion}
Next, Sec.~\ref{sec:imagesequences} is discussed. Topics like interaction types, spatio-temporal information modeling, slice length/location and prospects for future research directions are covered.

\textbf{Type of interaction.} Two works presented in this section are related to humans interacting either with a virtual agent~\cite{celiktutan2014continuous} or with small groups of people~\cite{aran2013cross}. Hypothetically, one could consider some kind of interaction exists when people talk to a camera, as in~\cite{biel2012facetube,Gurpinar:ECCV:2016,aran2013cross, Ventura:CVPRW2017,Edine17}. According to~\cite{aran2013cross}, people talk to the camera in \textit{vlogs} as if they were talking to other people. When some kind of interaction is considered, classes of features that encode specific aspects of social interactions can be exploited, such as visual activity, facial expressions or body/head motion.

\textbf{Spatio-temporal information.} A preliminary analysis suggests that the inclusion of spatio-temporal information placed first impression analysis on a new level (compared to still images), with a wider range of applications. Continuous predictions demonstrated to be a research line still little explored. This may be due to the challenging and complex task of generating accurate labels over time for a huge amount of data, in particular when deep learning based methods are considered, which from our knowledge, have not been employed in this context yet. Celiktutan and Gunes~\cite{celiktutan2014continuous} pioneered continuous predictions in first impressions. However, they used a small dataset composed of 30 video recordings captured from 10 subjects. Moreover, their continuous prediction can be interpreted as a frame-based regressor where each frame is treated independently. Thus, the dynamics of the scene are not completely explored. In~\cite{Edine17,Gurpinar:ECCV:2016}, short video clips are globally represented with statistics computed for the sequence of frames. Even though such approach does not treat the frames independently, it still does not consider the temporal evolution of the data. In~\cite{teijeiro2015your}, statistics of facial expression outputs are characterized as dynamic signals over brief observation windows, which can be considered a step forward to dynamically analyze first impression on the temporal dimension. 

Temporal information has been also exploited through relatively simple motion pattern representations~\cite{aran2013cross,Edine17}, which can have a positive impact on the outcomes. However, considered motion-based approaches are still quite limited. Very few works (presented in Sec.~\ref{sec:audiovisual}) proposed to model spatio-temporal dynamics using more powerful and advanced techniques such as 3D-CNN~\cite{Parteccv161} (to model local temporal patterns) or Long Short-Term Memory (LSTM) \cite{Celiktutan:TAC2017,Parteccv161}. Therefore, the influence (either in prediction or data labeling), benefits, as well as appropriate ways to model temporal information, are still open questions on the topic, as briefly discussed next (and later in Sec.~\ref{audiovisual:discussion}, w.r.t. \textit{slice length/location}) and in Sec.~\ref{discussion:multimodal} (\textit{future directions}).

\textbf{Slice length/location.} The predictive power of facial expressions, depending on the duration and relative position of specific \textit{vlog} segments, is analyzed in~\cite{teijeiro2015your}. Results suggest that viewers' impressions are better predicted by features computed at the beginning of each video, corroborating with the idea that first impressions are built from short interactions~\cite{Todorov:2017:book}. Nevertheless, authors reported that further research is needed to confirm their hypothesis, as well as to verify if the same effect is observed for different nonverbal sources, and whether the optimal duration and position of the slices are the same for each data type. 

\textbf{Future directions.} This study revealed that the number of methods developed for image sequences is significantly smaller if compared to those developed for audiovisual category (Sec.~\ref{sec:audiovisual}), which may be related to improvements obtained by the inclusion of complementary information or to the different applications being considered. Few works from the ones described in this section (e.g.,\cite{Gurpinar:ECCV:2016,celiktutan2014continuous}) have been extended to consider audiovisual information~\cite{Kaya17,BUNKU:ICPR:2016,Celiktutan:TAC2017}, emphasizing the benefits of including acoustic features to the pipeline. Nevertheless, in general, works are not exploiting the full benefits of temporal information. A feature representation that can keep the temporal evolution of the data, such as Dynamic Image Networks~\cite{Bilen:TPAMI2017} used in action recognition, should be considered in future research.

Although great advancements have been reported by deep learning based approaches~\cite{Gurpinar:ECCV:2016,Ventura:CVPRW2017}, they are often perceived as \textit{black-box} techniques, i.e., they are able to effectively model very complex problems, but cannot be easily interpreted nor their predictions can be explained. Because of this, explainability and interpretability should deserve special attention. 
In fact, the interest on these topics are evidenced by the organization of dedicated events, such as thematic workshops~\cite{WHI2017,NIPS:WKSyM2017} and challenges~\cite{Escalante:IJCNN:2017,Escalante:2018:arXiv:IJCV}. However, this kind of research is just on its infancy.

\subsection{Audiovisual trait prediction}\label{sec:audiovisual}
In this section, we review works using both acoustic (nonverbal) and visual features to perform automatic personality perception. Works are further classified based on the type of interaction (with/without), as they may use datasets, features and methodologies developed for different purposes. Interactive approaches, for instance, can exploit particular behaviours usually found in interactive scenarios (e.g., attention received while speaking), while non-interactive methods are generally focused on the individual.

\subsubsection{Interactive approaches}

Aran and Gatica-Perez~\cite{aran2013one}  
addressed personality perception during small group interactions using a subset of the ELEA corpus~\cite{sanchez2012nonverbal}. Thus, personality impressions needed to be collected, as the dataset did not provide them. The inference task is addressed using Ridge Regression combined with statistics computed from the given video segments (e.g., average speaking turn, \textit{prosodic} features and visual activity). For a comprehensive review of nonverbal cues applied to small group settings we refer the reader to~\cite{gatica2009automatic}.

Focusing on feature representation for personality and social impressions, Okada et al.~\cite{okada2015personality} proposed a co-occurrence event mining framework for \textit{multiparty} and multimodal interactions. According to the authors, the use of co-occurrence patterns between modalities yields two main advantages: (i) it can improve the inference accuracy of the trait value based on richer feature set and (ii) discover key context patterns linking personality traits. In their work, speech utterances, body motion and gaze are represented as time-series binary data, and co-occurrence patterns are defined as multimodal events overlapped in time. Then, co-occurrence events are detected through clustering before a final inference using Ridge Regression and linear SVM.

Staiano et al.~\cite{staiano2011automatic} focused on feature selection to model the dynamics of \textit{personality states} in a meeting scenario. Personality state 
refer to a specific behavioral episode wherein a person behaves as more or less introvert/extrovert, neurotic or
open to experience, etc. It is also referred as \textit{situational cues}~\cite{Fleeson:2001}. According to the authors, the problem with ``traditional approaches'' is that they assume a direct and
stable relationship between, e.g., being extravert and acting
extravertedly (speaking loudly, being talkative, etc). However, \textit{``on the
contrary, extraverts can sometimes be silent and reflexive, while
introverts can at times exhibit extraverted behaviors. Similarly,
people prone to neuroticism do not always exhibit anxious
behavior, while agreeable people can sometimes be aggressive''}~\cite{staiano2011automatic}. In their work, several low-level (acoustic) and high-level features (attention given/received in the form of head pose, gaze and voice activity) are combined with different machine learning approaches.

More recently, \c{C}eliktutan and Gunes~\cite{Celiktutan:TAC2017} addressed how personality impressions fluctuate with time and \textit{situational contexts}. First, audio-visual features are extracted (e.g., face/head/body movements, geometric and hybrid features). Then, a Bidirectional-LSTM Network is employed to model the temporal relationships between the continuously generated annotations and extracted features. Finally, a decision-level fusion is performed to combine the outputs of the audio and the visual regression models. Nevertheless, their study required a database (a subset of the SEMAINE corpus~\cite{mckeown2012semaine}) annotated in a time-continuous manner.

Interested on the differences in \textit{situational context} affecting trait perceptions and labelling, Joshi et al.~\cite{joshi2014automatic} analyzed thin slices ($14$ sec long) of behavioral data during HCI settings. Authors analyzed (i) the  differences between the perceived traits (Big-Five and social impressions) during audio-visual and visual-only observations; (ii) the deviation in the perception when there is a change of \textit{situational context}; (iii) and the change in the perception
marked by an external observer when the same individual interacts with different virtual characters (exhibiting specific emotional and social attributes). To take into account errors induced by subjective biases, they proposed a framework which encapsulates a weighted model based on linear SVR, and low-level visual features computed over the face region.

Bremner et al.~\cite{Bremner:HRI:2016} investigated how robot mediation affects the way the Big-Five personality traits of the operator are perceived. Results showed that (i) observers utilize robot appearance cues along with operator vocal cues to make their judgments; (ii) operators' gestures reproduced on the robot aid personality judgments, and (iii) that personality perception through robot mediation is highly operator-dependent. 
Extending~\cite{Bremner:HRI:2016}, \c{C}eliktutan et al.~\cite{Celiktutan:ROMAN:2016} showed that apparent personality classification from nonverbal cues works better than from audio-only (except for \textit{agreeableness}), and that facial activity and head pose together with audio and arm gestures play an important role in conveying specific personality traits in a telepresence context.

\subsubsection{Non-interactive settings}

In general, works falling in this category are those exploiting conversational videos, self-presentations or video resumes.

Biel and Gatica-Perez~\cite{biel2011you,biel2013youtube}  pioneered personality impressions in \textit{vlogs} under the perspective of audiovisual behavioral analysis. In~\cite{biel2011you}, they studied the use of nonverbal cues as descriptors of vloggers’ behavior. 
Later~\cite{biel2013youtube}, they addressed the  problem as a regression task, where features extracted from audio (speaking activity and prosodic), video (looking activity, pose and overall motion) and co-occurrence events (e.g., looking-while-speaking) were combined with SVR to infer apparent personality. In both works, the analyses are performed on thin \textit{vlog} slices (1-minute). An extensive review discussing both verbal and nonverbal aspects of \textit{vlogger} behaviors is presented in~\cite{Biel:THESIS:2013}.

Nguyen and Gatica-Perez~\cite{nguyen2016hirability} analyzed the formation of job-related first impressions in conversational video resumes. In fact, job recommendation systems based on the visual analysis of nonverbal human behavior started to receive a lot of attention from the past few years~\cite{nguyen2014hire, Nguyen:epfl:thesis:2015, Nguyen:ICMI:2013}, being social impressions the focus of most works. According to~\cite{nguyen2016hirability}, online video resumes represent an opportunity to study the formation of first impressions in an employment context at a scale never attempted before. In their work, the linear relationships between nonverbal behavior and the organizational constructs of hirability and apparent personality (Big-Five) are examined. Different regression methods are analyzed for the prediction of personality and hirability impressions from audio (speaking activity and prosody) and visual cues (proximity, face events and visual motion). Results suggest that combining feature groups strongly improves accuracy performance, reinforcing the benefits of including complementary information to the pipeline. More recently, Gatica-Perez et al.~\cite{Gatica-Perez:2018:ICMUM} addressed the recognition of \textit{personal state} and trait impressions in a longitudinal study using behavioral data of \textit{vloggers} who posted on YouTube for a period between three and six years. The dataset is composed of a small number of participants, and results do not show any significant temporal trend related to personality.

Following the recent advancements of CNNs, G{\"{u}}{\c{c}}l{\"{u}}t{\"{u}}rk et al.~\cite{guccluturk2016deep} presented an audiovisual Deep Residual Network (trained end-to-end) for apparent personality trait recognition. The network does not require any feature engineering or visual analysis such as face/landmark detection and alignment or facial expression recognition. Auditory and visual streams are merged in an
audiovisual stream, which comprises a fully-connected layer. At the training/test stage, the fully-connected layer outputs five continuous prediction values corresponding to each trait for the given input video clip. Their work won the third place in the ChaLearn First Impressions Challenge~\cite{lopez2016chalearn} (1\textsuperscript{st} round), whereas~\cite{Parteccv161} and~\cite{NJULambda} achieved the second and first place, respectively. The work~\cite{guccluturk2016deep} was extended in~\cite{Gucluturk:TAC:2017} to consider verbal content, and to predict an ``invitation to job interview'' variable. In~\cite{Parteccv161}, two end-to-end trainable deep learning architectures are proposed to recognize personality impressions. The networks have two branches, one for encoding audio and the other for visual features. The first model is formulated as a Volumetric (3D) CNN, while the second one is formulate as an LSTM based network, to learn temporal patterns in the audio-visual channels. Both models concatenate statistics of certain acoustic properties (obtained from non-overlapping partitions) and visual data (from segmented faces, after landmark detection) in a later stage.  

In order to capture rich information from both visual and audio modality, Zhang et al.~\cite{NJULambda,Wei:TAC:2017} proposed a Deep Bimodal Regression framework. They modified the traditional CNN to exploit important visual cues (introducing what they called Descriptor Aggregation Networks), and built a linear regressor for the audio modality. To combine complementary information from the two modalities, they ensembled predicted regression scores by both early and late fusion. G\"urpinar et al.\cite{BUNKU:ICPR:2016} extended their previous work~\cite{Gurpinar:ECCV:2016} (briefly described in Sec.~\ref{sec:imagesequences}) with the inclusion of other visual descriptors, acoustic features and weighted score level fusion strategy, and ranked in the first place of the ChaLearn First Impressions Challenge~\cite{icpr_contest} (2\textsuperscript{nd} round).

\subsubsection{Discussion}\label{audiovisual:discussion}
A general discussion about Sec.~\ref{sec:audiovisual} is presented next, covering current limitations, slice length/location, co-occurrence event mining, \textit{personality states}, job recommendation systems and prospect for future directions of research.

\textbf{Current limitations.} The ELEA~\cite{sanchez2012nonverbal} dataset, employed in~\cite{aran2013one}, was captured in a very specific and controlled environment, and developed for analyzing emergent leaderships. From the 40 recorded meetings, only 27 have both audio and video (for more details, see Table~\ref{tab:datasets}). From the point of view of deep learning based approaches, which are dominating different lines of research in social/affective computing, small datasets have limited application. The \textit{Mission Survival Task} corpus~\cite{mana2007multimodal}, employed in~\cite{staiano2011automatic}, according to the authors, is not currently available. The databases used in~\cite{Finnerty:ICMI:2016,Naim:FG:2015}, because of the privacy-sensitive content of the interviews or data protection laws~\cite{nguyen2016hirability}, are not publicly available. As recently reported in~\cite{Celiktutan:TAC2017}, most of the results found in the literature are not directly comparable to each other, as different evaluation protocols are employed. 

\textbf{Slice length/location.} According to~\cite{aran2013one}, external observers usually make their impressions based on thin slices selected from video samples. Thus, the decision of which part of the video (from the whole sequence) will be analysed is a common requirement to be addressed, and in most of the cases, it is done empirically. Slice length/location analysis is also studied in the context of social impressions from nonverbal visual analysis. Although having a different goal, it has strong overlap with personality perception studies as both areas are centered on the analysis of human behavior. For instance, Lepri et al.~\cite{lepri2010employing,Lepri:TAC2012} observed that classification accuracy (of \textit{extraversion} trait) is affected by the size of the slice when addressing \textit{real} personality recognition (i.e., classification performance increases with slice length up to a certain amount of time, and then it slightly decreases). In~\cite{Nguyen:epfl:thesis:2015},  manual ``scene/slice segmentation'' was performed to analyze different segments of a job interview in the context of social impressions. Results show that any slice clearly stood out in terms of predictive validity, i.e., all slices yielded comparable results. As stated in~\cite{Nguyen:epfl:thesis:2015}, which is shared by personality impressions studies, one of the challenges in thin slice research is the amount of temporal support necessary for each behavioral feature to be predictive of the outcome of the full interaction. In other words, some cues require to be aggregated over a longer period than others. According to their study, and to our review, no metric assessing the necessary amount of temporal support for a given feature exists, neither in social impressions nor in personality perception.
Thus, whether thin slice impressions would generalize to the whole video for the prediction task is still an open question. Question such as \textit{``is that possible to automatically select the best slice of the whole clip that best describes first impressions? Will it generalize to the whole video?''} could be a subject for future research in the field.

\textbf{Co-occurrence event mining.} It demonstrated to be very effective~\cite{okada2015personality,biel2013youtube,nguyen2016hirability} as an alternative to exploit complementary information associated to body language~\cite{Noroozi:arXiv2018}, such as \textit{looking while speaking} or \textit{attention received while silent}. According to~\cite{Huang:CVPRW2016}, the computer vision community has reached a point when it can start considering high-level reasoning tasks such as the ``communicative intents'' of images.  

\textbf{\textit{Personality states} or \textit{situational context}.} Have strong impact in personality perception, either when interactions are considered or not~\cite{staiano2011automatic,Fleeson:2001,Celiktutan:TAC2017,joshi2014automatic,Gatica-Perez:2018:ICMUM}. 
Although first impressions have been studied from different perspectives in the past few years, \textit{situational cues} did not receive enough attention from the computer vision community up to now. The great complexity and subjectivity related to this topic, along with the existing dataset limitations, could be a possible explanation for that. One can imagine, for instance, how challenge can be the task of defining a new dataset on this topic, either from different contexts (e.g., at work, in a party, during a job interview, etc) or even during the same context but at different time intervals~\cite{Celiktutan:TAC2017}. Although it can be considered a challenging task, future researches on this topic would have strong impact on the whole research area.

\textbf{Job recommendation.} The question of why a particular individual receives a positive (or negative) evaluation based on first impressions analysis deserves special attention from the research community, either if personality or hirability (social) impressions are considered. Note that a close link between the constructs of personality and hirability  exist~\cite{batrinca2011please}. Automatic job recommendation systems can be very subjective, and might have strong influence in our lives once they become common. Recent studies~\cite{Escalante:2018:arXiv:IJCV}, including those submitted to a workshop organized by the ChaLearn group~\cite{Escalante:IJCNN:2017}, sought to address such question. 

\textbf{Future directions.} Despite its limitations, verbal content~\cite{Biel:THESIS:2013}, combined with nonverbal visual data, is a potential direction to be taken to advance the state-of-the-art in automatic personality perception (discussed in Sec.~\ref{discussion:multimodal}).

Taking into account the particular case of job recommendation systems, the differences across job types have not been investigated in computing~\cite{nguyen2016hirability}, even though have already been addressed in psychological studies. For instance, the expected behavior for a person applying for a sales position may differ from someone looking for an engineering position. Combining differences across job types with personality trait analysis could be used to make job recommendation systems more transparent and inclusive.

Regarding the recently proposed CNN based models for automatic personality perception~\cite{guccluturk2016deep,Parteccv161,NJULambda,Ventura:CVPRW2017}, we observed that there is still a long venue to be explored. The top three winner methods~\cite{NJULambda, Parteccv161, guccluturk2016deep} submitted to the ChaLearn First Impression Challenge~\cite{lopez2016chalearn} obtained very similar overall performances (i.e., $0.913$, $0.912$ and $0.911$, respectively) even though presenting different solutions, suggesting that proposed architectures may be exploiting complementary features~\cite{Escalante:2018:arXiv:IJCV}, which could be combined somehow to improve overall accuracy. Moreover, deep neural networks are currently one of the most promising candidates to tackle the challenges of multimodal data fusion~\cite{Parteccv161,NJULambda,Wei:TAC:2017,BUNKU:ICPR:2016} and multi-task solutions in first impressions.

\subsection{Multimodal trait prediction}\label{sec:multimodal}
This section reviews works using multimodal data for automatic personality perception, i.e., in addition to the audio-visual cues, they may exploit verbal content, depth information or use data acquired by more specialized devices.

Biel et al.~\cite{Biel:ICMI:2013} addressed  personality impressions of \textit{vloggers} using Linguistic Inquiry and Word Count (LIWC) and N-grams analysis. While the focus of their work is on what \textit{vloggers} say, few experiments fusing verbal and nonverbal content have been performed. Verbal content is also exploited in~\cite{Biel:THESIS:2013}. In this case, the work focuses on nonverbal content analysis. Both works~\cite{Biel:ICMI:2013,Biel:THESIS:2013} use manual transcripts of \textit{vlogs} to verify (in an error-free setting) the ability of verbal content for the prediction of personality impressions (Big-Five). The feasibility of building a fully automatic framework were investigated using Automatic Speech Recognition (ASR). However, errors caused by the ASR system significantly decreased the performances. 

Ch\'{a}vez-Mart\'{\i}nez et al.~\cite{Chavez-Martinez:2015:HAM:2836041.2836051} considered the inference of mood and personality impressions (Big-Five) of \textit{vloggers}, from verbal content (i.e., categorizing word counts into linguistic categories, obtained from manual transcriptions) and nonverbal audio-visual cues  (e.g., pitch, speaking rate, body motion and facial expression). High-level facial features are considered through the concept of compound facial expressions. The inference task is then addressed using a multi-label classifier. Results suggest that the combination of mood and trait labels improved overall performance in comparison with the mood-only and trait-only experiments.

Using a logistic regression model, Sarkar et al.~\cite{Sarkar:MMWCPR:2014} combined audiovisual (pitch, speech and movement analysis), verbal content (unigram bag of words and statistics from the transcriptions), demographic (gender) and sentiment features (e.g., positive/negative sentiment scores of the verbal content) for apparent personality trait (Big-Five) classification. Results show that different personality traits are better predicted using different combinations of features. Alam and Riccardi~\cite{Alam:WCPR:2014} reached a similar conclusion when addressing personality impression using the same dataset (Youtube \textit{vlog}~\cite{biel2013youtube}), i.e., the performance of each trait varies for different feature sets. In their work, linguistic, psycholinguistic and emotional features extracted from the transcripts are analyzed, in addition to the audio-visual features provided with the dataset. Similarly as in~\cite{Sarkar:MMWCPR:2014,Alam:WCPR:2014}, Farnadi et al.~\cite{Farnadi:2014:MRA:2659522.2659526} combined audiovisual and several text-based features with different multivariate regression techniques. The main differences among the above solutions~\cite{Sarkar:MMWCPR:2014,Alam:WCPR:2014,Farnadi:2014:MRA:2659522.2659526} are in relation to  verbal features and the way the problem was modeled, as audio-visual features were provided with the dataset (released in the WCPR2014 Challenge~\cite{Celli:2014:WCP:2647868.2647870}).

Srivastava et al.~\cite{srivastava2012don} exploited audio, visual and lexical
features to predict the Big-Five Inventory-10 answers, from which personality trait scores can be obtained. Audiovisual and verbal cues are combined for recognizing emotions, and used to learn a linear regression model based on the proposed Sparse and Low-rank Transformation (SLoT). A dataset composed of short movie clips (4-7 sec long each), manually labeled with BFI-10 answers and personality impressions is used. Several high level tasks are performed (e.g., face/emotion expression recognition, tracking, scene change detection) as the dataset can show multiple people and cuts, which makes the study even harder. Moreover, as dialogs are extracted from the movie's subtitles, audiovisual information might not be well synchronized with the text.

More recently, G{\"{u}}{\c{c}}l{\"{u}}t{\"{u}}rk et al.~\cite{Gucluturk:TAC:2017} extended their previous work~\cite{guccluturk2016deep} to consider verbal content (extracted from audio transcripts provided with the data) as well as to infer hirability scores~\cite{Escalante:IJCNN:2017,Escalante:2018:arXiv:IJCV}. Different modalities are analysed, including audio-only, visual-only, language-only, audiovisual, and a combination of audiovisual and language in a late fusion strategy. Results show that the best performance is obtained though the fusion of all data modalities. 

Exploiting the concept of \textit{group engagement}, Salam et al.~\cite{Salam:2017} investigated how personality impressions of participants can be used together with robot's personality to predict the engagement state of each participant in a triadic Human-Human-Robot Interaction setting. Nonverbal visual cues of individuals (e.g., body activity, appearance and visual focus of attention) and interpersonal features (e.g, relative distances, attention given/received) captured from RGB-D data are employed. However, several high level tasks have to be addressed for feature extraction, such as ROI/group detection, body/head/face detection, skeleton joints, as well as robot detection. Authors also propose to use extroverted/introverted robots in the experiments to vary the context of the interactions. 

Finnerty and collaborators~\cite{Finnerty:ICMI:2016} studied whether first impressions of stress are equivalent to physiological measurements of electrodermal activity (EDA) in a context of job interviews. The outcomes of job interviews are then analyzed based on features extracted from multiple data modalities (EDA, audio and visual). Even though focusing on stress impressions, they presented preliminary analysis on the relationship among \textit{real} and apparent personality, and stress impressions, briefly discussed in Sec.~\ref{APR:section}. 

\subsubsection{Discussion}\label{discussion:multimodal}
In this section, we discuss different aspects of complementary information, recurrent problems on the field, as well as prospects for future research directions related to Sec.~\ref{sec:multimodal}. Then, we close the section with a summary of recent trend methodologies with respect to all data modalities.

\textbf{Complementary information.} Whether focusing on verbal~\cite{Biel:ICMI:2013} or nonverbal content analysis~\cite{Biel:THESIS:2013}, this study revealed that overall improvements of each personality trait are obtained when different cues are employed~\cite{Biel:ICMI:2013,Sarkar:MMWCPR:2014,Alam:WCPR:2014}, as well as that they can be further improved when different features are combined/fused~\cite{Gucluturk:TAC:2017}. For instance,  improvements were obtained in~\cite{Chavez-Martinez:2015:HAM:2836041.2836051} by combining mood and trait labels.  In~\cite{Biel:ICMI:2013}, \textit{extraversion} was better predicted using nonverbal content, whereas \textit{agreeableness}, \textit{neuroticism} and \textit{conscientiousness} when verbal cues were used. 
Nevertheless, according to the revised literature, verbal content analysis has some limitations: 1) most existing works exploiting verbal content are based on manual transcriptions of the audio channel (which imposes a great barrier to applicability); 2) automatic speech recognition methods are still not so accurate to capture verbal content without introducing noise/errors to the pipeline; and 3), it is language dependent, i.e., verbal content analysis from people of different spoken languages might require different treatments. 
Thus, according to our study, there is not a single set of features that maximizes accuracy performance for all personality traits. Verbal content, voice, facial expressions, gestures and poses (i.e., body language), among many other features, are potential sources to code/decode personality, and can complement each other in different ways. 

\textbf{Recurrent problems.} The use of controlled environments~\cite{Salam:2017} or specialized sensors~\cite{Finnerty:ICMI:2016} imposes a limitation to applicability, which can be even more limited if the study is based on private~\cite{Salam:2017, srivastava2012don} or customized datasets~\cite{Finnerty:ICMI:2016,Chavez-Martinez:2015:HAM:2836041.2836051} composed of small number of participants~\cite{Salam:2017}.

\textbf{Future directions.} The two-stage approach presented in ~\cite{srivastava2012don} is motivated by the fact that the relationship between features and personality traits is generally difficult to describe through a simple linear model. Authors attempted to learn a model for predicting answers to BFI-10 questionnaire from features, which is like a \textit{mid-level} step in understanding the semantic hierarchy from features to personality traits. Results indicate that \textit{features to answers} followed by \textit{answers to personality scores} can achieve superior performance than \textit{features to personality scores}, even though in a preliminary study. As far as we know, no other work on the field tried to address the problem using such two-stage methodology, which could receive special attention in future researches.

It is well known that deep learning is making a revolution with respect to almost all research domains related to visual computing (i.e., compared to traditional machine learning and standard computer vision approaches based on hand-crafted features). The architecture presented in~\cite{guccluturk2016deep,Gucluturk:TAC:2017} was the only approach, among the top ranking approaches in the ChaLearn First Impression Challenge~\cite{lopez2016chalearn}, which
did not rely neither on pretrained models nor on feature engineering, which makes it particularly appealing since it does not require making any assumptions regarding the important features for the task at hand. Authors also evaluated the changes in performance for the audio and visual models as a function of exposure time (i.e., \textit{slice length}), which still is an open question on the field. Results suggest that there is enough information about personality in a single frame~\cite{Gucluturk:TAC:2017}, as evidenced in~\cite{Todorov:2017:book} when studying the bases of personality judgments (``\textit{first impressions are built from a glimpse as brief as $100$ms}''). Nevertheless, the same reasoning does not hold for the auditory modality, especially for very short auditory clips. Note that, frame selection (i.e., which frame from the whole sequence will be analyzed), not addressed in~\cite{Gucluturk:TAC:2017}, have not been properly addressed yet in spatio-temporal based methods for automatic personality perception (i.e., the standard approach is to analyse uniformly distributed samples over the set of frames).

Although single images can carry meaningful information about the personality of an individual, we envisage future research directions in multimodal approaches (which hold for almost all other categories of the proposed taxonomy) should contemplate end-to-end trainable models, multi-task scenarios (different tasks are jointly trained), taking benefit of the evolution of the data on the spatio-temporal domain, possibly benefiting from transfer learning (e.g., cross-domain) and semi-supervised approaches (using partially annotated data from different datasets).

Finally, our study also revealed recent trend methodologies, summarized in Table~\ref{tab:trendmodels}, where deep architectures predominate. Nevertheless, there is sill not a standard way to represent multimodal information, i.e., hand-crafted, deep features and raw data are combined in different ways. Therefore, some models are able to obtain high accuracy performance without any engineered feature (at least with respect to the visual channel), which can be considered a break trough in automatic personality perception. It can be also observed that most of these trend methodologies are benefiting from transfer learning through the use of pretrained models, which increases complexity and number of parameters, but can be considered a step forward to enhance generalization to different contexts and scenarios. The design of standard evaluation protocols, databases and challenges (discussed in Sec.~\ref{challenges:section}), facilitates the comparison of different approaches. For instance, if we consider the databases \cite{lopez2016chalearn} and \cite{Escalante:IJCNN:2017} (shown in Table~\ref{tab:trendmodels}), they are composed by the same audio-visual data (see Table~\ref{tab:datasets} for more details) and we can directly compare works evaluated on them. However, there is still a large venue to be explored, as some \textit{key-points} affecting personality perception can be considered at their early stages of research (e.g., the influence of face attributes and context, spatio-temporal modeling, explainability, etc).

\begin{table*}[htbp]
\centering
\caption{Recent trend methodologies in apparent personality trait analysis. Cat.: still image (\textbf{SI}), image sequence (\textbf{IS}), audio-visual (\textbf{AV}) and multimodal~(\textbf{M}). \textit{Pretrained} models employed for feature extraction or pre-processing (audio end video resampling are not considered) are reported. Average \textit{Score} (and metric) reported for a particular \textit{Database} is presented.  Legend: \textbf{LFAV}: late fusion of acoustic and visual cues; \textbf{FLDA}: face/landmark detection and alignment; \textbf{HC}: hand-crafted; \textbf{DF}: deep feature; \textbf{RD}: raw-data; \textbf{acc}: accuracy.}
\label{tab:trendmodels}
\setlength\tabcolsep{1.5pt}
\begin{tabular}{|C{0.6cm}|C{0.6cm}|C{4.5cm}|C{1.4cm}|C{1.9cm}|C{2.4cm}|C{1.3cm}|C{1.0cm}|C{3.4cm}|}
\hline
\textbf{Ref} & \textbf{Cat.} & \textbf{Model} &  \textbf{Pretrained} & \textbf{Pre-processing} & \textbf{Features} & \textbf{Database} & \textbf{Score} & \textbf{Key-points$^*$}\\ \hline
\cite{Dhall:HBU:2016} & SI &  KPLS multivariate regression & VGG16 & FLDA & HD and DF & custom from \cite{Liu:AAAI:2016} & 0.37 (RMSE) & scene descriptor \\ \hline
\cite{Gurpinar:ECCV:2016} & IS & KELM regressor (late fusion of scene and emotion features) &  VGG19; VGG-Face & FLDA & HD and DF & \cite{lopez2016chalearn} & 0.9094
 (acc) & scene descriptor; facial expression \\ \hline
\cite{Ventura:CVPRW2017} & IS & Deep Bimodal Regression &  VGG-Face & face detection & HD and DF & \cite{Escalante:IJCNN:2017} & 0.912
 (acc) &  Action Unit analysis; explainability \\ \hline
\cite{Celiktutan:TAC2017} & AV & BLSTM with a decision level fusion of audio and visual cues &    - & FLDA; upper body detection  & HD & custom from \cite{mckeown2012semaine}& 0.51 (MSE) & continuous domain; varying context \\ \hline
\cite{Parteccv161} & AV & 3D-CNN and LSTM bi-modal deep
networks (LFAV) &   - & FLDA & RD (video); HC (audio) & \cite{Escalante:IJCNN:2017} & 0.9121
(acc) & end-to-end; spatio-temporal modeling \\ \hline
\cite{Wei:TAC:2017} & AV & Deep Bimodal Regression  (early \& LFAV); linear regression (audio) &  VGG-Face; ResNet & - & RD (video); HC (audio) & \cite{lopez2016chalearn} & 0.9130
(acc) & ensemble of multiple models; explainability \\ \hline
\cite{Gucluturk:TAC:2017} & M & RNN (LFAV); Ridge regression (LFAV and verbal content) &   - & - & RD (video, audio); HC (verbal) & \cite{Escalante:IJCNN:2017} & 0.9118 (acc) & end-to-end; explainability; verbal content \\ \hline
\multicolumn{9} {l} {$^*$ \textit{Key-points} are particular aspects we consider relevant to be further exploited.}
\end{tabular}
\end{table*}

%
%
\subsection{\textit{Real} and apparent personality trait analysis}\label{APR:section}
In this section, we present relevant works developed for automatic personality recognition, due to their relevance and similarity to the topic of this survey, and briefly discuss the very few existing works analysing both \textit{real} and apparent personality traits. Note that, if we ignore the way labels (used to train machine learning algorithms for the classification/regression task) are obtained (i.e., from external observers, in the case of personality perception, or from self-report questionnaires, in \textit{real} personality trait analysis), the task of inferring personality from visual, audio-visual, or multimodal data could be considered the same, either if \textit{real} or apparent personality is considered. However, it is important to emphasise that ``apparent personality recognition'' is not a replacement for ``real personality recognition'', as personality perception does not necessarily relate to the actual trait(s) of the person.

\subsubsection{Automatic Personality Recognition}

Similarly as the case of personality perception, personality recognition have been addressed in the literature using different data modalities, i.e., \textit{still images}~\cite{Ferwerda:ICMM:2016,Liu:AAAI:2016,Celli:2014:API:2647868.2654977}, \textit{image sequences}~\cite{Celiktutan:ROMAN:2015,subramanian2013relationship}, \textit{audiovisual} (with~\cite{Fang:ICMI:2016,Pianesi:2008:MRP:1452392.1452404,lepri2009modeling,kalimeri2010causal,lepri2010employing,Lepri:TAC2012,Celiktutan:FG:2015,batrinca2012multimodal,Nguyen:ICMI:2013} or without~\cite{batrinca2011please,batrinca2011multimodal} interactions) and \textit{multimodal}~\cite{rahbar2015predicting,Anzalone2017,farnadi16,Abadi:FG:2015}.

Taking the popularity of social networks into account, Ferwerda et al.~\cite{Ferwerda:ICMM:2016}  proposed to infer \textit{real} personality from the way users manipulate their pictures on Instagram. However, the work is basically limited to color information analysis. Liu et al.~\cite{Liu:AAAI:2016} addressed how Twitter profile images vary with the personality of users. Although profile images from over 66,000 users are used, personality traits were estimated based on their tweets. In~\cite{Celli:2014:API:2647868.2654977}, they analysed the Big-Five traits and interaction styles from Facebook profile images. However, without explicitly analysing human faces, as some images even contain one single person. 

Subramanian et al.~\cite{subramanian2013relationship} show that social attention patterns computed from target's position and head pose during social interactions are excellent predictors of \textit{extraversion} and \textit{neuroticism}. In~\cite{Celiktutan:ROMAN:2015}, the impact of personality during Human-Robot Interactions (HRI) is analysed based on nonverbal cues extracted from a first-person perspective, as well as from their relationships with participants' self-reported personalities and interaction experience. Linear SVR is employed to predict personality traits from gaze direction, attention and head movements while interacting with either an ``extroverted'' or ``introverted'' robot.

Fang et al.~\cite{Fang:ICMI:2016} combined Ridge Regression with audio-visual features to address Big-Five personality recognition and social impressions during small group interactions. In a similar context, speaking time and visual attention is exploited in~\cite{lepri2010employing,Lepri:TAC2012} to predict \textit{extraversion}.  In~\cite{lepri2010employing}, they consider the attention an individual receive/gives from/to the group members. In~\cite{Lepri:TAC2012}, authors differentiate the amount of attention given/received while the person is speaking. Both works also address the impact of \textit{slice length} on the classification. In a meeting scenario, Pianesi et al.~\cite{Pianesi:2008:MRP:1452392.1452404} addressed \textit{extraversion} and \textit{Locus of Control} classification using SVM and audio-visual features (e.g. audio signal statistics and \textit{Motion History Images}) extracted from 1-min videos. Later~\cite{lepri2009modeling}, the problem was addressed as a regression task. Inference of \textit{Extraversion} and \textit{Locus of Control} are also proposed in~\cite{kalimeri2010causal} using a Bayesian Networks that explicitly incorporate hypotheses about the relationships among personality, actual behavior of the target and \textit{situational aspects}.

Batrinca et al.\cite{batrinca2012multimodal} employed 2-5 min videos to recognize personality traits during HCI, combining audio-visual cues and feature selection with SVM. In their work, the computer interacts with individuals using different levels of collaborations, to elicit the manifestation of different personality traits.
The work was extended in~\cite{Batrinca:TM:2016} to consider Human-Human Interactions (HHI). In~\cite{batrinca2011please}, authors combined  nonverbal visual features with Naive Bayes and SVM to predict Big-Five traits in a similar monologue setting than \textit{vlogging}~\cite{biel2011you}. 
The study~\cite{batrinca2011please} was extended in~\cite{batrinca2011multimodal} to automatically extract few additional visual features. In both works, the highest accuracy were obtained when classifying \textit{conscientiousness}. As related by the authors, the request of introducing themselves in front of a camera, apparently activated the subjects' \textit{conscientiousness} dispositions. 

Nguyen et al.~\cite{Nguyen:ICMI:2013} extend~\cite{nguyen2014hire} to predict the Big-Five traits in addition to hirability impressions, focusing on postures and gestures (extracted from a mixture of manual annotations and automated methods) as well as co-occurrence events. Rahbar et al.~\cite{rahbar2015predicting} addressed \textit{extraversion} recognition during HRI, taking into account the first thin slices of the interaction. Multimodal features extracted from depth images (e.g., motion and human-robot distance) are used to train a Logistic Regression Classifier. The work was extended in~\cite{Anzalone2017} to consider new features and in depth analysis. Fernadi et al.~\cite{farnadi16} compared different personality recognition methods and investigated the possibility of cross-learning from different social media, i.e. Facebook, Twitter and YouTube. However, disregarding the analysis performed on the YouTube \textit{vlog}~\cite{biel2013youtube} dataset, any visual-based analysis was performed in relation to the other sources. In~\cite{Abadi:FG:2015}, they studied the relation between Big-Five traits and implicit responses of people to affective content (i.e., emotional videos), combining features obtained from \textit{electroencephalogram}, peripheral physiological signals and facial landmark trajectories with a linear regression model.

In summary, the very few existing works developed for automatic personality recognition are: \textbf{1)} mainly based on hand-crafted features, classic machine learning approaches and single-task scenarios, without modelling multiple visual human cues for an accurate representation neither exploiting temporal dynamics of the data; \textbf{2)} analysed on small sized datasets without standard evaluation protocols. In consequence, there is no generalization guarantee to different target populations/scenarios; and last but not least \textbf{3)} none of the existing works regressing personality traits from audio-visual data analyse sources of bias to further correlate \textit{real} and apparent personality.

\subsubsection{Joint analysis of real and apparent personality}

Preliminary results on the relationship between \textit{real} and apparent personality traits have been reported in the literature~\cite{Wolffhechel:2014,Mairesse:2007,Bremner:HRI:2016,Finnerty:ICMI:2016,Celiktutan:2017:TAC} with limited outcomes. In the work of Wolffhechel et al.~\cite{Wolffhechel:2014}, any connection was found between participants self-reported personality traits and the scores they gave to others. According to their study, a single facial picture may lack information for evaluating diverse traits, i.e., \textit{``a viewer will miss additional cues for gathering a more complete first impression and will therefore focus overly on facial expressions instead''}.
In~\cite{Mairesse:2007}, no visual information is used (just audio and transcripts). Furthermore, a small dataset was used and results show low generalization capability of the proposed model. Bremner et al.~\cite{Bremner:HRI:2016} analysed audio-visual and audio-only information on a small dataset of 20 participants (and 5 observers) captured from a controlled environment, with no generalization guarantee to different target populations. They observed that judge’s ratings bear a significant relation to the target's self-ratings only for the \textit{extraversion} trait.

Although restricted to job interviews, preliminary results on the relationship among impressions of stress and the Big-Five traits (\textit{real} and apparent) are reported in~\cite{Finnerty:ICMI:2016}. 
With respect to \textit{real} personality, they observed that \textit{Openness to experience} and \textit{Conscientiousness} traits were negatively correlated with stress impressions. For traits as perceived by others, \textit{Conscientiousness} was negatively correlated with stress impressions while \textit{Neuroticism} was positively correlated.

More recently, Celiktutan et al.~\cite{Celiktutan:2017:TAC} presented a multimodal database to study personality simultaneously in HHI and HRI scenarios, and its relationship with engagement. Note that, differently from existing approaches in personality computing, personality impressions were provided by acquaintance people, which may explain why baseline results show that trends in personality classification performance remained the same with respect to the self and acquaintance labels. Moreover, results are limited to an analysis conducted with a small number of participants. 

In~\cite{Guntuku:2015}, authors addressed \textit{real} and apparent personality traits recognition. However, in addition to presenting some insights on which cues contribute to each personality type, any joint analysis was performed. Fang et al.~\cite{Fang:ICMI:2016} addressed the recognition of \textit{real} personality traits and social impressions without making an in depth correlation analysis about both domains. In~\cite{aran2013one}, personality impressions were collected for the same dataset used in~\cite{Fang:ICMI:2016}. However, still without correlating both personality types.

This study revealed that state-of-the-art in visual-based personality computing are neither focusing on understanding human biases that influence personality perception nor trying to automatically (and accurately) regress the real personality from perception mechanisms. This is mainly because of the high complexity on accurately modelling humans in visual data, as well as the high subjectivity of the topic (involving a large set of possible sources of bias), remaining a largely unexplored area.

\subsection{What features give better results?}\label{features:section}

As discussed in previous sections, there is not a standard set or modality of features that works better for any type of data, database or personality trait. Different solutions have been proposed over the past years based on distinct evaluation protocols, which prevent the above question to be properly addressed. However, we present in Table~\ref{tab:feat:correlation} a list of mid-level features and semantic attributes highly correlated with Big-Five traits, reported by state-of-the-art works on personality computing. We expect the information summarized in Table~\ref{tab:feat:correlation} can be used to inspire future researches to advance the state-of-the-art on the field, in particular when studying new strategies to improve the recognition performance of traits that are currently difficult to be recognized (as discussed in Sec.~\ref{recognitionaccuracy:section}).

While some agreement can be observed in Table~\ref{tab:feat:correlation} (in most of the cases) in relation to particular sets of attributes, traits and personality type, few minor inconsistencies can also be noted, reinforcing the difficulty of addressing the above question. This is the case of ``positive emotions and smiling expressions'' with respect to \textit{openess} and  \textit{real} personality (reported to have, in different studies, negative and positive correlation). It must be emphasized that the studies performed by the different works presented in Table~\ref{tab:feat:correlation} may be limited to analysis performed on small datasets, composed of small number of participants and without standard evaluation protocols, which do not guarantee generalization to different scenarios and contexts. In the case of apparent personality, they are also influenced by subjective bias from people who labelled the data, number of annotators, among other variables. Nevertheless, disregarding these limitations, it is possible to observe some agreement with reported literature in psychology, such as that extroverted people usually show smiling expressions and speak louder, as well as that more conscientious people tend to preserve their privacy (e.g., by not sharing images from private locations on social media), for example. 

Evidences found in the literature show that \textit{extraversion} is the trait showing the largest activity cue utilization~\cite{biel2012facetube} (reflected in Table~\ref{tab:feat:correlation}), as well as it is the trait typically easier to judge~\cite{biel2011you,biel2013youtube}. Our study also revealed that, for the case of personality perception, \textit{extraversion} is also the trait recognized with higher accuracy (shown in Fig.~\ref{traits:fig}).

\begin{table*}[htbp]
\centering
\caption{Reported mid-level features and semantic attributes, highly correlated with Big-Five traits, reported by state-of-the-art in personality computing.}
\label{tab:feat:correlation}
\setlength\tabcolsep{1.5pt}
\begin{tabular}{|C{3.0cm}|C{4.4cm}|C{0.6cm}|C{4.0cm}|C{5.5cm}|}
\hline
\multicolumn{2}{|c|}{\textbf{Negative correlation}}  & \multirow{2}{*}{\textbf{Trait}} & \multicolumn{2}{c|}{\textbf{Positive correlation}}\\ \cline{1-2} \cline{4-5}
\textbf{\textit{Real}} & \textbf{Apparent} &  & \textbf{\textit{Real}} & \textbf{Apparent} \\ \hline
Positive emotions and smiling expressions \cite{Liu:AAAI:2016}; stress impression~\cite{Finnerty:ICMI:2016} &
Negative emotions (anger, disgust) \cite{Biel:THESIS:2013, biel2012facetube, teijeiro2015your}; verbal content associated to negative emotions \cite{Biel:ICMI:2013}; long eye contact \cite{biel2013youtube} and frontal face event duration \cite{nguyen2016hirability} &
\textbf{O} &
Positive emotions and smiling expressions [45]; body motion and speaking activity during collaborative task \cite{Batrinca:TM:2016} &
Smiling expressions, joy \cite{teijeiro2015your}; speaking time \cite{biel2011you}; body motion \cite{Biel:THESIS:2013, biel2013youtube, nguyen2016hirability}; hirability impressions and eye contact \cite{nguyen2016hirability}; verbal content associated to leisure activities \cite{Biel:ICMI:2013} \\ \hline
Private location \cite{Guntuku:2015}; negative mood \cite{Liu:AAAI:2016}; speech conflict with others \cite{Fang:ICMI:2016}; stress impression \cite{Finnerty:ICMI:2016} &
Negative emotions/valence (sad, anger) \cite{teijeiro2015your, Biel:ICMI:2013}; body activity \cite{biel2011you}; long eye contact \cite{Biel:THESIS:2013, biel2013youtube}; stress impression \cite{Finnerty:ICMI:2016} &
\textbf{C} &
Smiling expressions \cite{batrinca2011please}, positive mood and valence (joy) \cite{Liu:AAAI:2016}; eye contact \cite{Celli:2014:API:2647868.2654977}; speaking activity and body motion during collaborative task \cite{Batrinca:TM:2016} &
Smiling expressions, joy and contempt \cite{teijeiro2015your}, frontal pose \cite{Dhall:HBU:2016}; speaking time \cite{biel2011you}; eye contact, low body motion activity, looking-while-speaking \cite{Biel:THESIS:2013, biel2013youtube}; verbal content associated to occupation and achievements \cite{Biel:ICMI:2013} \\ \hline
Low deviation in received attention during interaction \cite{subramanian2013relationship};  visual attention given to the rest of the group~\cite{kalimeri2010causal} &
Neutral or negative emotions/valence (anger, disgust, contempt) \cite{Biel:THESIS:2013, teijeiro2015your, Pianesi:2008:MRP:1452392.1452404, Chavez-Martinez:2015:HAM:2836041.2836051,Gatica-Perez:2018:ICMUM}; pressed lips \cite{Guntuku:2015}; low speech activity \cite{Gatica-Perez:2018:ICMUM} during group interaction \cite{okada2015personality}; give/receive attention while silent \cite{Lepri:TAC2012}; speech turns \cite{Biel:THESIS:2013, biel2013youtube, biel2011you}; long eye contact \cite{biel2013youtube, nguyen2016hirability} &
\textbf{E} &
Positive emotions \cite{Liu:AAAI:2016}; smiling expressions \cite{Celli:2014:API:2647868.2654977}; body motion \cite{Batrinca:TM:2016, batrinca2012multimodal}; attention received and speaking time \cite{kalimeri2010causal}; engagement during interaction~\cite{Salam:2017} &
Positive emotions/valence (joy)\cite{Guntuku:2015, Biel:THESIS:2013, biel2012facetube, Chavez-Martinez:2015:HAM:2836041.2836051} and smiling expressions\cite{biel2012facetube, teijeiro2015your}; funny\cite{Gatica-Perez:2018:ICMUM}; body motion\cite{Biel:THESIS:2013, biel2013youtube, nguyen2016hirability, biel2011you, okada2015personality}; give/receive attention while speaking\cite{biel2013youtube, okada2015personality, Lepri:TAC2012, aran2013one}; speaking time\cite{Biel:THESIS:2013, biel2013youtube, biel2011you} and louder\cite{biel2013youtube}; eye contact\cite{nguyen2016hirability}; verbal content related to interpersonal interactions and sexuality\cite{Biel:ICMI:2013}; hirability impression\cite{nguyen2016hirability, Nguyen:ICMI:2013} \\ \hline
Negative emotion expressions \cite{Liu:AAAI:2016}; talking turns in group interactions \cite{Fang:ICMI:2016}; &
Negative emotions/valence (anger and disgust) \cite{Biel:THESIS:2013, biel2012facetube, teijeiro2015your, Chavez-Martinez:2015:HAM:2836041.2836051}; verbal content associated to negative emotions, sexuality and religion \cite{Biel:ICMI:2013} &
\textbf{A} &
Positive emotions, smiling expressions and joy \cite{Liu:AAAI:2016}; body motion during collaborative task \cite{Batrinca:TM:2016}, and long speaking duration \cite{batrinca2011please} &
Positive emotions/valence (joy) and smiling expression \cite{Biel:THESIS:2013, biel2012facetube, teijeiro2015your,Gatica-Perez:2018:ICMUM}; eye contact \cite{Guntuku:2015}; facial attractiveness \cite{joshi2014automatic}; verbal content associated to positive emotions \cite{Biel:ICMI:2013}  \\ \hline
Positive emotions \cite{Liu:AAAI:2016}; smiling expressions \cite{Celli:2014:API:2647868.2654977}; long speaking duration during collaborative task \cite{Batrinca:TM:2016} &
Smiling expressions, joy \cite{teijeiro2015your}; face visibility \cite{Guntuku:2015}; facial attractiveness \cite{joshi2014automatic}; looking-while-speaking \cite{Biel:THESIS:2013, biel2013youtube}; hirability impression \cite{nguyen2016hirability}; aggreableness \cite{Gatica-Perez:2018:ICMUM}&
\textbf{N} &
Negative emotions or lack of emotion expression \cite{Liu:AAAI:2016}; social profile images without faces \cite{Liu:AAAI:2016, Celli:2014:API:2647868.2654977}; low body motion activity during interactions \cite{subramanian2013relationship} &
Negative emotions/valence (anger) \cite{Chavez-Martinez:2015:HAM:2836041.2836051,Gatica-Perez:2018:ICMUM}; verbal content associated to negative words and negative emotional words \cite{Biel:ICMI:2013}; duckface \cite{Guntuku:2015}; stress impression \cite{Finnerty:ICMI:2016} \\ \hline
\end{tabular}
\end{table*}

\subsection{Which traits are easily recognized?}\label{recognitionaccuracy:section}

\begin{figure}[htbp]
	\centering
	\includegraphics[height=2.5cm]{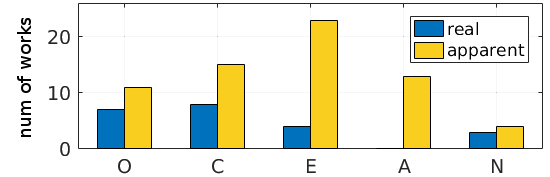}
	\caption{Distribution of Big-Five personality traits ``easily'' recognized.}
	\label{traits:fig}
\end{figure}

To address the above question, we analysed the results reported by reviewed works with respect to the Big-Five model, i.e., \{O, C, E, A, N\}. In total, $11$ and $33$ works addressing \textit{real} and apparent personality recognition, respectively, have been analysed. Works which did not report results for all Big-Five traits have not been considered. For each work, we retrieved the two traits recognized with highest accuracy. Fig.~\ref{traits:fig} shows the obtained distribution. Different observations can be taken form Fig.~\ref{traits:fig}: \textbf{1)} the ``ranking'' of traits on the two distributions are different, i.e., \textbf{2)} regarding \textit{real} trait estimation, ``C'' and ``O'' are the traits recognized with highest accuracy, whereas \textbf{3)} ``E'' and ``C'' are the best recognized traits in personality perception; \textbf{4)} clearly, ``N'' is challenging for both types of work; \textbf{5)} surprisingly, ``A'', which is usually recognized with satisfactory accuracy in personality perception, is the most difficult trait to be recognized when \textit{real} personality is considered.

It must be emphasized that Fig.~\ref{traits:fig} was generated based on few works evaluated on different datasets and protocols. Thus, the analysis presented in this section can only provide a more general view around the above question, and further studies are needed to confirm our observations.

\section{Trait recognition challenges}\label{challenges:section}

Academic competitions/challenges are an excellent way to quickly advance the state-of-the-art in a particular field. Organizers of challenges formulate a problem, provide data, evaluation metrics, evaluation platform and forums for the dissemination of results. Within the computer vision, pattern recognition and multimedia information processing communities\footnote{Please, note that inferring \textit{real} personality traits from text is a topic that has been studied intensively by the NLP community (e.g.,~\cite{Shared13}).} several challenges related to personality analysis have been proposed. These are summarized in Table~\ref{sota:challenges}.

\begin{table*}[th]
	\centering
	\caption{Challenges on apparent personality analysis. }
	\label{sota:challenges}
    \begin{tabular}{|c|c|c|c|c|c|c|c|}
\hline
\textbf{Challenge} & \textbf{Year} & \textbf{Dataset} & \textbf{Task} & \textbf{Samples}& \textbf{Event} & \textbf{Winner} & \textbf{Overview} \\ \hline
\multirow{3}{*}{ChaLearn LAP} & 2017 & First impressions v2  & Invite to interview &10,000& CVPR/IJCNN & \cite{Kaya17,Edine17} & \cite{Escalante:IJCNN:2017}\\ \cline{2-8}
& 2016 & First impressions & P.  traits (Big-5)&10,000& ECCV & \cite{NJULambda}  & \cite{lopez2016chalearn}
 \\ \cline{2-8}
& 2016 & First impressions  & P. traits (Big-5)&10,000&  ICPR & \cite{BUNKU:ICPR:2016} &\cite{icpr_contest} \\\hline
MAPTRAITS&2014&Semaine& P. traits (Big-5)&44&ICMI&\cite{Kaya:2014:CMP:2668024.2668025,Sidorov:2014:ARP:2668024.2668028}&\cite{Celiktutan:2014:MFA:2668024.2668026}\\ \hline
WCPR&2014&YouTube \textit{vlog}& P. traits (Big-5)&404&MM&\cite{Alam:WCPR:2014}&\cite{Celli:2014:WCP:2647868.2647870}\\ \hline
Speaker Trait&2012&I-ST&P. traits (Big-5)&640&Interspeech&\cite{IvanovInterspeech12}&\cite{SullerInterspeech12}\\\hline
\end{tabular}
\end{table*}

The first challenge on apparent personality analysis was that organized at Interspeech in 2012~\cite{SullerInterspeech12}. However, the challenge focused only in the audio modality. Interestingly, organizers of such competition found that participants had difficulty at improving the baseline, and solutions lacked \emph{creativity}, hence motivating further research in terms of the  feature extraction and modeling .

The MAPTRAITS~\cite{Celiktutan:2014:MFA:2668024.2668026} and WCPR~\cite{Celli:2014:WCP:2647868.2647870} challenges organized in 2014 were the first ones involving both video and audio modalities, focusing on personality perception. Two tracks were launched in  the MAPTRAITS challenge: a continuous estimation of personality traits through time  and the recognition of traits from entire clips of video (discrete). Organizers found that participants barely obtained comparable performance as the baselines. Where the best results were obtained with audio-visual and visual only baselines, for the continuous and discrete tracks, respectively. Hence highlighting the importance of the visual modality. The WCPR challenge also comprised two tracks: one focusing on the use  multimodal information and the other using only text information. The conclusions from this competition were that the problem was too hard, but better results were obtained with multimodal information than when only text was used. 
In both challenges, the number of available samples were quite small (only 44 samples were used in~\cite{Celiktutan:2014:MFA:2668024.2668026} and 404 in~\cite{Celli:2014:WCP:2647868.2647870}), which may be the cause for the low recognition performance. Still, these were the first efforts on the usage of multimodal information for personality analysis.  

More recently,  ChaLearn\footnote{\url{http://chalearnlap.cvc.uab.es}} organized two rounds of a challenge on  personality perception from audiovisual data~\cite{lopez2016chalearn,icpr_contest}. A new dataset comprising realistic videos annotated by AMT workers was used. To the best of our knowledge, this is the largest dataset available so far for apparent personality analysis (10K samples). The challenge focused on trait recognition in short clips taken from YouTube. The winning methods were based on deep learning~\cite{NJULambda,BUNKU:ICPR:2016}. In fact, most participants of the contest adopted deep learning methods (e.g.,~\cite{Parteccv161,guccluturk2016deep}).  The best performance was achieved by solutions that incorporated both audio and visual cues. For these competitions, participants succeeded at improving the performance of the baseline, achieving recognition performance above 90\% of average accuracy.  The main conclusion from these competitions was that accurately recognizing personality impressions from short videos is feasible, motivating further research in this topic.  

Results from the latter challenge motivated a new competition in a closely related topic involving personality traits, the so called Job Candidate Screening Coopetition~\cite{Escalante:IJCNN:2017}. In this challenge, an extended version of the ChaLearn First Impression dataset~\cite{lopez2016chalearn} was considered, where the extension consisted of the new variable to be predicted and manual transcriptions of audio in the videos. Participants had to predict a variable indicating whether the person in a video would be invited or not to a job interview (round 1). In addition, participants had to develop an interface explaining their recommendations (round 2). For the latter task, participants relied on apparent personality trait predictions.  Organizers concluded that the \textit{``invite-for-interview''} variable could be predicted with high accuracy, and that developing explainable models is a critical aspect, yet there is still a large room for improvement in this aspect.

\begin{table*}[htbp]
\caption{Available datasets used in personality computing, centered on the visual analysis of humans, from a vision-based perspective.}
\label{tab:datasets}
\setlength\tabcolsep{1.5pt}
\begin{tabular}{|C{2.0cm}|C{0.6cm}|C{6.0cm}|C{4.0cm}|C{3.0cm}|C{1.8cm}|}
\hline
\textbf{Dataset} & \textbf{Year} & \textbf{Short description} & \textbf{Focus} & \textbf{Labels} & \textbf{Used in}\\ \hline
MHHRI~\cite{Celiktutan:2017:TAC}, \textit{Multimodal} & 2017 & 12 interaction sessions (\texttildelow4h) captured with egocentric cameras, depth and bio-sensors, 18 participants, \textit{controlled environment} & Personality and engagement during HHI and HCI &   Self/acquaintance-assessed Big-Five, and engagement &  \cite{Celiktutan:ROMAN:2015,Salam:2017} \\ \hline
ChaLearn First Impression v2 \cite{Escalante:IJCNN:2017}, \textit{Multimodal} & 2017 & Extended version of~\cite{lopez2016chalearn}, with the inclusion of hirability impressions and audio transcripts & Apparent personality trait and hirability impressions &  Big-Five impressions, job interview variable and transcripts  &  \cite{Escalante:2018:arXiv:IJCV,Ventura:CVPRW2017,Gucluturk:TAC:2017,Kaya17,Edine17} \\ \hline
ChaLearn First Impression~\cite{lopez2016chalearn}, \textit{Audiovisual} & 2016 &  10K short videos:  \texttildelow15sec each, collected from 2762 YouTube users, 1280x720 of size, RGB, 30fps, \textit{uncontrolled environment}   & Apparent personality trait analysis (no interaction - single person talking to a camera) &  Big-Five impressions &  \cite{guccluturk2016deep,Parteccv161,NJULambda,Wei:TAC:2017,Gurpinar:ECCV:2016,BUNKU:ICPR:2016,icpr_contest} \\ \hline
SEMAINE \cite{mckeown2012semaine}, \textit{Multimodal} & 2012 & 959 conversations: \texttildelow5min each, 150 participants, 780x580 of size, 49.979fps, RGB and gray, frontal and profile view, \textit{controlled environment} & Face-to-face (interactive) conversations with sensitive artificial listener agents &  Metadata$^*$, transcripts, 5 affective dimensions and 27 associated categories$^\dagger$ &  \cite{celiktutan2014continuous,Celiktutan:FG:2015,Celiktutan:2014:MFA:2668024.2668026,joshi2014automatic,Celiktutan:TAC2017,Kaya:2014:CMP:2668024.2668025,Sidorov:2014:ARP:2668024.2668028}
\\ \hline
Emergent LEAder (ELEA)~\cite{sanchez2012nonverbal}, \textit{Audiovisual} & 2012 & 40 meetings: \texttildelow15min each, 27 having both audio and video, composed of 3 or 4 members, 148 participants; 6 static (25fps) and 2 portable (30fps) cameras, \textit{controlled environment} & Small group interactions and emergent leadership (winter survival task) & Metadata$^*$, Big-Five (self-report)$^\dagger$ and social impressions  &  \cite{aran2013one,okada2015personality,Fang:ICMI:2016} \\ \hline
YouTube \textit{vlog} \cite{biel2010vlogcast,biel2010voices,biel2013youtube},  \textit{Audiovisual} & 2011 & 442 \textit{vlogs}: \texttildelow$1$min each, 1 video per participant, \textit{uncontrolled environment}  & Conversa\-tional
\textit{vlogs} \cite{biel2010vlogcast,biel2010voices} and apparent personality trait analysis~\cite{biel2013youtube}  &  Metadata$^*$ and Big-Five impressions & \cite{biel2011you,biel2012facetube,biel2013youtube,Biel:ICMI:2013,teijeiro2015your,aran2013cross,Biel:THESIS:2013,Sarkar:MMWCPR:2014,Alam:WCPR:2014,Farnadi:2014:MRA:2659522.2659526}\\ \hline
\multicolumn{6} {l} {$^*$ \textit{Metadata} can be gender, age, number of ``likes'' on social media, presence of laughs, Facial Action Units, etc, and vary for each dataset.}\\
\multicolumn{6} {l} {$^\dagger$ Labels for personality perception are annotated by each independent work, as they are not provided with the database.}
\end{tabular}
\end{table*}

From these challenges, several lessons can be learned. First, a major difficulty for organizers of the first competitions was the scarcity of data. From that competitions, participants could barely improve the performance of the baseline and there was not too much diversity in terms of the type of solution. Larger datasets, as those provided in the more recent challenges, together with more powerful modeling techniques that can leverage such amounts of data (e.g., deep learning methods), could be the key component for the participants to succeed at improving baselines and obtaining outstanding performance. Table~\ref{tab:datasets} shows available datasets on the topic, from a vision-based perspective, as well as additional details about labels, modality, etc. Secondly, the inclusion of multimodal information, e.g., as opposed to using audio or text only, increased the range of possible methodologies and information that could be used to predict personality traits. Finally,  the initial competitions were not organized for consecutive years (except those organized by ChaLearn). We think continuity is a key aspect for the success of any academic challenge. 

It is important to emphasize that despite there are not too many challenges on personality analysis, the progress that previous competitions have caused is remarkable. In addition, there are several related challenges  that deserve to be mentioned\footnote{We consider these challenges are related because both are associated to the social signal processing field.}. For instance, the EmotiW~\cite{Dhall:2016:EVG:2993148.2997638}  and Avec~\cite{Avec16} challenge series, that have been run for several years and whose focus is on emotion recognition, depression detection, mood classification, and multimodal data processing. 
Clearly, progress in these related fields is having an impact on new challenges targeting personality traits exclusively.

\section{Discussion}\label{conslusion:section}

%
In this section, we provide a final discussion about the research topic. First, we comment few and relevant observed aspects, and lessons learned, at different modalities of the proposed taxonomy. Then, we summarize and discuss the changes in terms of applications, features and limitations, when temporal information started being used. Next, we discuss accuracy performance and its relation to subjectivity, as well as the importance of dataset definition as rich resources to advance the research. Finally, we comment current deep learning technologies applied on first impression analysis, expected outcomes and applications.


\textbf{Particularities of the different modalities.} From this study, several lessons can be learned. It revealed that still images based approaches mostly focus on geometric and/or appearance facial features, using low-level or mid-level/semantic attributes to drive the recognition of personality traits. Most works within this category use \textit{ad-hoc} datasets, making the comparison with competitive approaches a big challenge. Techniques developed for image sequences usually include higher level features and analysis, such as facial emotion expressions, co-occurrence event mining, head/body motion, in addition to the ones used for still images. When temporal information is available, the great majority of works tend to compute functional statistics over time or treat each frame independently, omitting large spatio-temporal interactions. We envision future studies exploiting the spatio-temporal dependencies in (audio)visual (or multimodal) data to be an essential line of research. Possible studies in this direction may focus on new temporal deep learning models, such as using 3D convolutions (to consider local motion patterns), or based on temporal models such as Recurrent Neural Networks (RNN)-LSTM, which is able to model large spatio-temporal interactions.

\textbf{Beyond still images.} The use of temporal information introduced the problem of defining the slice duration and location. Even though addressed in some works, these questions remain open in all related modalities. Some other issues appeared when audiovisual approaches got in focus, such as \textit{situational contexts} or \textit{personality states}, which are extremely important points, as they contribute to increase the complexity and subjectivity in first impression studies. 

In addition to  visual information,  methods for the analysis of personality in videos have relied on other modalities of data. For instance, most of the works reviewed in this survey fall within the audiovisual category, including low-level acoustic features (pitch, intensity, frequencies) as well as descriptors of speaking activity, turns, pauses, looking-while-speaking (co-occurrence event), etc. This is because nonverbal audio modality has proven to carry information that can be highly  correlated with personality. In the same line, lexical analysis from the audio transcriptions has proven to be very useful, this is not surprising as automatic detection of personality from text has been widely studied. 

\textbf{Performance \textit{vs} subjectivity.} A considerable number of works show that the performance of each trait varies from different feature sets. This study also revealed: 1) the ranking of best recognized traits varies with respect to reviewed works, although some tendencies have been observed; 2) the best feature set/modality, or ranking of best recognized traits, might also change for the same work from one dataset to another due to subjectivity and complexity of the task.

\textbf{Public datasets as valuable resources.} A major problem in personality computing in the past has been  the lack of unified public datasets for allowing the accurate evaluation of methodologies for personality recognition and perception. Our review revealed that the construction of resources can be a valuable contribution. In fact, nowadays there are a few datasets available already, some of them generated in the context of academic challenges. Nevertheless, the design of new datasets and challenges will speed up the progress in the field at fast rates. We envision new, large and public datasets, considering a large amount of heterogeneous population, and exploiting the following topics could define new research directions in the next few years: 1) \textit{situational contexts} or \textit{personality states} in more realistic scenarios; 2) continuous predictions; 3) joint analysis of \textit{real} and apparent personality; 4) observer \textit{vs.} observed analysis, in the context of first impression data labeling and subjective bias analysis. Note that these topics are highly correlated, and could be somehow tackled together, being a richer source for research on the field. They could also potentially benefit from 5) an increased comprehensive personality profile (which could also consider physical and mental health~\cite{Ware:1992}, cognitive abilities~\cite{Raven:2003}, implicit bias~\cite{Baron:2019}, among other attributes).

\textbf{The revolution of deep learning.} This study revealed that most works developed for automatic personality trait analysis (either if \textit{real} or apparent) are mainly based on hand-crafted features, standard machine learning approaches and single-task scenarios. Although few recent works are trained end-to-end, they do not integrate a comprehensive set of human visual cues, neither address human bias nor correlate \textit{real} and apparent personality. Nevertheless, CNNs are starting to be used in first impressions with very promising results, allowing the model to analyze not only a limited set of predefined features but the whole scene with contextual information, as well as facilitating advanced spatio-temporal modeling through the use of, e.g., RNN or 3D-CNN. Furthermore, CNNs can be considered nowadays on of the most promising candidates to meet the challenges of multimodal data fusion by virtue of its high capability in extracting powerful high-level features.

\textbf{Outcomes.} Recent and promising results on personality computing may encourage psychologists to get interested in machine learning approaches and to contribute to the field. Along these lines, a very promising venue for research has to do with the incorporation of prior knowledge in personality analysis models. This represents a new challenge for both psychology and  machine learning communities. Likewise, we believe that users of personality recognition methods would benefit from information on the decisions or recommendations made by an automatic system.  Thus, a quite promising venue for research is explainability and interpretability of personality recognition methods.  

\textbf{Applications.} This study revealed that automatic personality trait analysis is applied in a vast number of scenarios. Reviewed works are applied in social media, small groups, face-to-face or interface based interviews, HRI, HCI, \textit{vlogs}, video resumes, among others contexts. From a practical point of view, the wide range of potential applications related to automatic personality perception, whose limits have not been defined yet, can benefit health, affective interfaces, learning, business and leisure, among others. However, in order to be applied only for good causes, the new generation if intelligent systems provided with personality perception capabilities will need to be more effective, explainable, and inclusive, being able to ethically generalize to different cultural and social contexts, benefiting everyone, everywhere. We anticipate personality computing will become a hot topic in the next few years with a high impact in a wide number of applications and scenarios.


%



\ifCLASSOPTIONcompsoc
  \section*{Acknowledgments}
\else
  \section*{Acknowledgment}
\fi

This project has been partially supported by granted Spanish Ministry projects TIN2016-74946-P,  TIN2015-66951-C2-2-R and TIN2017-88515-C2-1-R. This work is partially supported by ICREA under the ICREA Academia programme. We thank ChaLearn Looking at People sponsors for their support, including Microsoft Research, Google, NVIDIA Corporation, Amazon, Facebook and Disney Research.

\ifCLASSOPTIONcaptionsoff
  \newpage
\fi



%


\bibliographystyle{IEEEtran}


%

\begin{IEEEbiography}[{\includegraphics[width=1in,height=1.25in,clip,keepaspectratio]{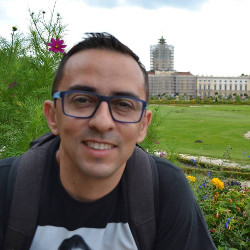}}]{Julio C. S. Jacques Junior} 
is a postdoctoral researcher at the Computer Science, Multimedia and Telecommunications department at Universitat Oberta de Catalunya (UOC), within the Scene Understanding and Artificial Intelligence (SUNAI) group. He also collaborates within the Computer Vision Center (CVC) and Human Pose Recovery and Behavior Analysis (HUPBA) group at Universitat Autonoma de Barcelona (UAB) and University of Barcelona (UB), as well as within ChaLearn Looking at People. 
\end{IEEEbiography}

\vskip -1.2\baselineskip plus -1fil

\begin{IEEEbiography}[{\includegraphics[width=1in,height=1.25in,clip,keepaspectratio]{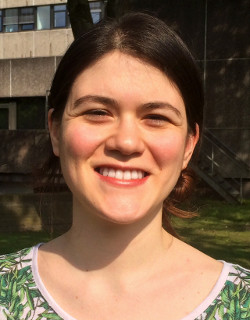}}]{Yağmur Güçlütürk} is an assistant professor 
at the Radboud University, Donders Institute for Brain, Cognition and Behaviour, Nijmegen, Netherlands, where she is working as a member of the Artificial Cognitive Systems Lab and the Neuronal Stimulation for Recovery of Function Consortium. Previously, she did her Ph.D. and M.Sc in Cognitive Neuroscience at the Radboud University, and B.IT in Artificial Intelligence at the Multimedia University. 
\end{IEEEbiography}

\vskip -1.2\baselineskip plus -1fil

\begin{IEEEbiography}
[{\includegraphics[width=1in,height=1.25in,clip,keepaspectratio]{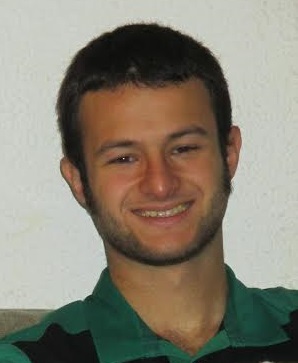}}]{Marc P\'{e}rez} obtained the B.S. in Computer Science and Mathematics at University of Barcelona. His research interests include computer vision and machine learning, with special interest in affective computing and self-driving cars.
\end{IEEEbiography}

\vskip -1.2\baselineskip plus -1fil

\begin{IEEEbiography}[{\includegraphics[width=1in,height=1.25in,clip,keepaspectratio]{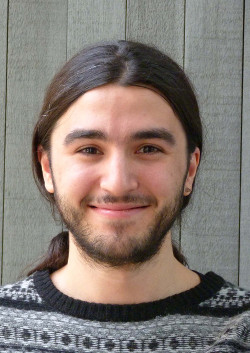}}]{Umut Güçlü} is an assistant professor at Radboud University, Donders Institute for Brain, Cognition and Behaviour, Nijmegen, Netherlands, and a member of the Artificial Cognitive Systems lab, where he works on combining deep learning and neural coding to systematically investigate the cognitive algorithms in vivo with neuroimaging.
He did his postdoctoral, doctoral and master's studies in cognitive neuroscience and bachelor's studies in artificial intelligence.
\end{IEEEbiography}

\vskip -1.2\baselineskip plus -1fil

\begin{IEEEbiography}[{\includegraphics[width=1in,height=1.25in,clip,keepaspectratio]{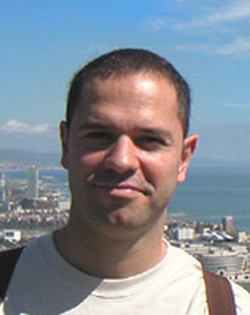}}]{Carlos Andujar}
 is associate professor at the Computer Science Department of the Universitat Politecnica de Catalunya - BarcelonaTech, and senior researcher at the Research Center for Visualization, Virtual Reality and Graphics Interaction, ViRVIG.
He received his PhD in Software Engineering in 1999 from UPC.
His research interests include 3D modeling, computer graphics, and virtual reality. 
He is currently the vicedirector of the CS department at UPC. 
\end{IEEEbiography}

\vskip -1.2\baselineskip plus -1fil

\begin{IEEEbiography}[{\includegraphics[width=1in,height=1.25in,clip,keepaspectratio]{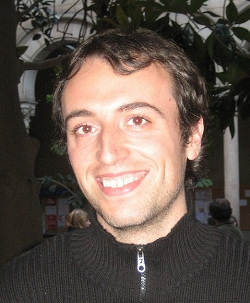}}]{Xavier Bar\'{o}}
is associate professor and researcher at the Computer Science, Multimedia and Telecommunications department at Universitat Oberta de Catalunya (UOC). From 2015, he is the director of the Computer Vision Master degree at UOC. He is co-founder of the SUNAI group, and collaborates within CVC at UAB as member of HUPBA group. He is interested on machine learning, evolutionary computation, and pattern recognition, specially their applications to the field of Looking at People.
\end{IEEEbiography}

\vskip -0.7\baselineskip plus -1fil

\begin{IEEEbiography}
[{\includegraphics[width=1in,height=1.25in,clip,keepaspectratio]{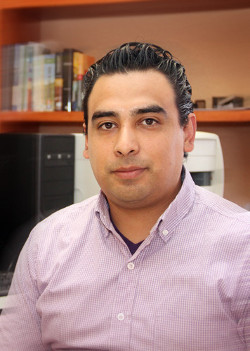}}]{Hugo Jair Escalante}  is researcher scientist at Instituto Nacional de Astrofisica, Optica y Electronica, INAOE, Mexico, 
and a director of ChaLearn, a nonprofit dedicated to organizing challenges, since 2011. He has been involved in the organization of several challenges in computer vision and automatic machine learning. He has served as coeditor of special issues in IJCV, PAMI, and TAC, and as as competition/area chair of venues like NeurIPS, PAKDD and IJCNN.  
\end{IEEEbiography}

\vskip -0.7\baselineskip plus -1fil

\begin{IEEEbiography}
[{\includegraphics[width=1in,height=1.25in,clip,keepaspectratio]{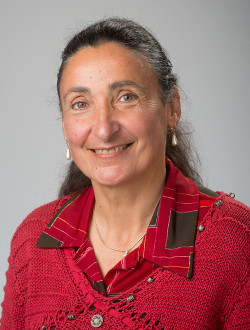}}] {Isabelle Guyon} is chaired professor in ``big data” at the Universit\'e Paris-Saclay, specialized in statistical data analysis, pattern recognition and machine learning (ML). Prior to joining Paris-Saclay she worked as an independent consultant and was a researcher at AT\&T Bell Laboratories, where she pioneered applications of neural networks to pen computer interfaces (with collaborators including Yann LeCun and Yoshua Bengio) and co-invented with Bernhard Boser and Vladimir Vapnik Support Vector Machines. She organizes challenges in ML since 2003 supported by the EU network Pascal2, NSF, and DARPA, with prizes sponsored by Microsoft, Google, Facebook, Amazon, Disney Research, and Texas Instrument. She is action editor of the Journal of ML Research.
\end{IEEEbiography}

\vskip -0.7\baselineskip plus -1fil

\begin{IEEEbiography}[{\includegraphics[width=1in,height=1.25in,clip,keepaspectratio]{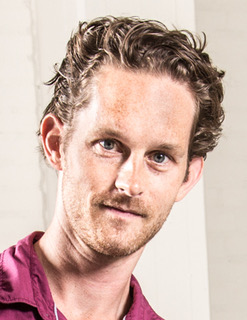}}]{Marcel A. J. van Gerven}
is professor of Artificial Cognitive Systems at Radboud University and  principal investigator in the Donders Institute for Brain, Cognition and Behaviour. His research focuses on brain-inspired computing, understanding the neural mechanisms underlying natural intelligence and the development of new ways to improve the interaction between humans and machines. He is head of the Artificial Intelligence department at Radboud University.\end{IEEEbiography}

\vskip -0.7\baselineskip plus -1fil

\begin{IEEEbiography}
[{\includegraphics[width=1in,height=1.25in,clip,keepaspectratio]{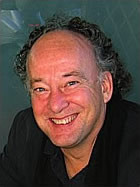}}]{Rob van Lier}
studied Experimental Psychology, and after that he took his PhD on the topic of human visual perception at Radboud University (Nijmegen, The Netherlands). Next, he spent a few postdoc years at the University of Leuven (Belgium) and returned to Nijmegen, based on a two-year grant from the Dutch National Science Foundation (NWO) and a 5-year grant from the Royal Netherlands Academy of Arts and Sciences (KNAW). Currently, he is an associate professor at the Radboud University and a principal investigator at the Donders Institute for Brain, Cognition and Behaviour, heading the ``Perception and Awareness'' research group.
\end{IEEEbiography}

\vskip -0.7\baselineskip plus -1fil

\begin{IEEEbiography}[{\includegraphics[width=1in,height=1.25in,clip,keepaspectratio]{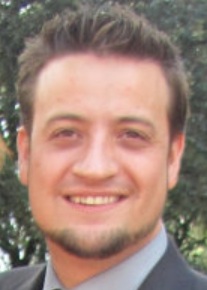}}]{Sergio Escalera} obtained the best P.h.D. award in 2018 at Computer Vision Center, UAB. He leads the Human Pose Recovery and Behavior Analysis Group (HUPBA). He is an associate professor at the Department of Mathematics and Informatics, Universitat de Barcelona (UB). He is also a member of the Computer Vision Center (CVC) at UAB. He is vice-president of ChaLearn Challenges in Machine Learning and chair of IAPR TC-12: Multimedia and visual information systems. He has been awarded with ICREA Academia. His research interests include the visual analysis of humans, with special interest in affective and personality computing.

 \end{IEEEbiography}






\end{document}